\documentclass[twoside]{article}

\usepackage[accepted]{aistats2020}
\usepackage[utf8]{inputenc} 
\usepackage[T1]{fontenc}    
\usepackage{hyperref}       
\usepackage{booktabs}       
\usepackage{amsfonts}       
\usepackage{nicefrac}       
\usepackage{microtype}      
\usepackage{subcaption}
\usepackage{cite}
\usepackage{answers}
\usepackage{setspace}
\usepackage{graphicx}
\usepackage{enumitem}
\usepackage{multicol}
\usepackage{mathrsfs}
\usepackage{amsmath,amsthm,amssymb}
\usepackage{algorithm}
\usepackage[noend]{algpseudocode}
\usepackage{color}
\usepackage{dsfont}

\DeclareMathOperator*{\argmax}{arg\,max}
\DeclareMathOperator*{\argmin}{arg\,min}

\renewcommand{\P}{\mathbb{P}}
\newcommand{\E}{\mathbb{E}}

\newcommand{\R}{\mathbb{R}}

\newcommand{\vx}{\boldsymbol{x}}
\newcommand{\vI}{\boldsymbol{I}}
\newcommand{\vK}{\boldsymbol{K}}
\newcommand{\vy}{\boldsymbol{y}}

\newcommand{\vk}{\boldsymbol{k}}
\newcommand{\vc}{\boldsymbol{c}}

\newcommand{\Ic}{\mathcal{I}}

\newcommand{\Ac}{\mathcal{A}}
\newcommand{\Sc}{\mathcal{S}}
\newcommand{\Fc}{\mathcal{F}}
\newcommand{\Hc}{\mathcal{H}}
\newcommand{\bone}{\mathds{1}}

\newcommand{\ty}{\tilde{y}}
\newcommand{\tmu}{\tilde{\mu}}

\newcommand{\ucb}{\mathrm{ucb}}
\newcommand{\lcb}{\mathrm{lcb}}
\newcommand{\ucbbar}{\overline{\mathrm{ucb}}}
\newcommand{\lcbbar}{\overline{\mathrm{lcb}}}

\newcommand{\ucbs}{\mathrm{ucb}^{(S)}_{t_S - 1}}
\newcommand{\lcbs}{\mathrm{lcb}^{(S)}_{t_S - 1}}
\newcommand{\ucbsbar}{\overline{\mathrm{ucb}}^{(S)}_{t_S - 1}}
\newcommand{\lcbsbar}{\overline{\mathrm{lcb}}^{(S)}_{t_S - 1}}

\newcommand{\ucbf}{\mathrm{ucb}^{(F)}_{t_F - 1}}
\newcommand{\lcbf}{\mathrm{lcb}^{(F)}_{t_F - 1}}
\newcommand{\ucbfbar}{\overline{\mathrm{ucb}}^{(F)}_{t_F - 1}}
\newcommand{\lcbfbar}{\overline{\mathrm{lcb}}^{(F)}_{t_F - 1}}

\newcommand{\ucbabar}{\overline{\mathrm{ucb}}^{(A)}_{t_A - 1}}

\newcommand{\ucbap}{{\mathrm{ucb}}^{(A)}_{t'_A - 1}}
\newcommand{\lcbap}{{\mathrm{lcb}}^{(A)}_{t'_A - 1}}
\newcommand{\ucblp}{{\mathrm{ucb}}^{(\ell)}_{t'_\ell - 1}}
\newcommand{\lcblp}{{\mathrm{lcb}}^{(\ell)}_{t'_\ell - 1}}

\newcommand{\lcbst}{\mathrm{lcb}^{(S)}_{t_S}}

\newcommand{\lcbstbar}{\overline{\mathrm{lcb}}^{(S)}_{t_S - 1}}

\newcommand{\ucbft}{\mathrm{ucb}^{(F)}_{t_F}}

\newcommand{\ucbftbar}{\overline{\mathrm{ucb}}^{(F)}_{t_F - 1}}

\newcommand{\ucbat}{\mathrm{ucb}^{(A)}_{t_A}}
\newcommand{\lcbat}{\mathrm{lcb}^{(A)}_{t_A}}
\newcommand{\ucbatbar}{\overline{\mathrm{ucb}}^{(A)}_{t_A - 1}}
\newcommand{\lcbatbar}{\overline{\mathrm{lcb}}^{(A)}_{t_A - 1}}
\newcommand{\ucbltbar}{\overline{\mathrm{ucb}}^{(\ell)}_{t_\ell - 1}}
\newcommand{\lcbltbar}{\overline{\mathrm{lcb}}^{(\ell)}_{t_\ell - 1}}

\newcommand{\ucbalg}{\mathrm{ucb}^{(A_1)}_{t}}
\newcommand{\lcbalg}{\mathrm{lcb}^{(A_1)}_{t}}

\newcommand{\isValid}{\mathrm{isValid}}

\newcommand{\gpucb}{\textsc{GP-UCB}}
\newcommand{\fsucb}{\textsc{Fast-Slow GP-UCB}}

\newtheorem{theorem}{Theorem}

\newtheorem{lemma}[theorem]{Lemma}

\begin{document}

%

%

\twocolumn[

\aistatstitle{Corruption-Tolerant Gaussian Process Bandit Optimization}

\aistatsauthor{ Ilija Bogunovic \And Andreas Krause \And  Jonathan Scarlett }
\aistatsaddress{ ETH Z\"urich \And  ETH Z\"urich \And National University of Singapore } ]

\begin{abstract}
    \vspace*{-2ex}
    We consider the problem of optimizing an unknown (typically non-convex) function with a bounded norm in some Reproducing Kernel Hilbert Space (RKHS), based on noisy bandit feedback.  We consider a novel variant of this problem in which the point evaluations are not only corrupted by random noise, but also adversarial corruptions.  We introduce an algorithm $\fsucb$ based on Gaussian process methods, randomized selection between two instances labeled ``fast'' (but non-robust) and ``slow'' (but robust), enlarged confidence bounds, and the principle of optimism under uncertainty.  We present a novel theoretical analysis upper bounding the cumulative regret in terms of the corruption level, the time horizon, and the underlying kernel, and we argue that certain dependencies cannot be improved. We observe that distinct algorithmic ideas are required depending on whether one is required to perform well in both the corrupted and non-corrupted settings, and whether the corruption level is known or not.
\end{abstract}

\vspace*{-3ex}
\section{Introduction}
\vspace*{-1ex}

Bandit optimization problems on large or continuous domains have far-reaching applications in modern machine learning and data science, including robotics \cite{lizotte2007automatic}, hyperparameter tuning \cite{snoek2012practical}, recommender systems \cite{vanchinathan2014explore}, environmental monitoring \cite{srinivas2009gaussian}, and more.  To make such problems tractable, one needs to exploit correlations between the rewards of ``similar'' actions.  In the {\em kernelized multi-armed bandit} (MAB) problem, this is done by utilizing smoothness in the form of a low function norm in some Reproducing Kernel Hilbert Space (RKHS), permitting the application of 
 Gaussian process (GP) methods \cite{srinivas2009gaussian,chowdhury17kernelized}.  See \cite[Ch.~6]{rasmussen2006gaussian} for an introduction to the connections between GPs and RKHS functions.

Key theoretical developments for the RKHS optimization problem have included both upper and lower bounds on the performance, measured via some notion of regret \cite{srinivas2009gaussian, chowdhury17kernelized,scarlett2017lower}.  The vast majority of these results have focused only on zero-mean additive noise in the point evaluations, and as a result, it is unclear to what extent the performance degrades under {\em adversarial corruptions}.  Such considerations are of significant interest under erratic or unpredictable sources of corruption, and particularly arise when the samples may be perturbed by a malicious adversary.  As we argue in Section \ref{sec:problem}, prominent algorithms such as GP-UCB \cite{srinivas2009gaussian} can be quite brittle in the face of such corruptions.

In this paper, we study the optimization of RKHS functions with both random noise and adversarial corruptions.  We propose a novel algorithm and regret analysis building on recently-proposed techniques for the finite-arm stochastic MAB setting \cite{lykouris2018stochastic}.  Specifically, we present a randomized algorithm $\fsucb$ based on randomly choosing between a ``fast'' non-robust instance, and a ``slow'' robust instance.  We bound the cumulative regret of $\fsucb$ in terms of the adversarial corruption level, time horizon, and underlying kernel.

The kernelized setting comes with highly non-trivial additional challenges compared to the finite-arm setting, primarily due to the infinite action space and correlations between their associated function values.  In particular, while correlations are undoubtedly beneficial in the non-corrupted setting (taking a given action permits learning something about similar actions), this benefit can lead to a hindrance in the corrupted setting:  An adversary that corrupts a given sample can potentially \emph{damage} our belief regarding {\em many} nearby function values.  Moving beyond independent arms was posed as a open problem in \cite[Sec.~5.3]{gupta2019better}.

\textbf{Related work on GP optimization.}
Numerous GP-based bandit optimization algorithms have been proposed in recent years~\cite{srinivas2009gaussian, hennig2012entropy, hernandez2014predictive, bogunovic2016truncated, wang2017max, shekhar2017gaussian, ru2017fast}. Beyond the standard setting, several important extensions have been considered, including multi-fidelity~\cite{bogunovic2016truncated,kandasamy2017multi,song2018general}, contextual and time-varying settings~\cite{krause2011contextual, valko2013finite, bogunovic2016time},
safety requirements~\cite{sui2015safe}, high-dimensional settings~\cite{djolonga2013high, kandasamy2015high, rolland2018high}, and many more.

Certain types of corruption-tolerant GP-based optimization algorithms have been explored previously, with the defining features including (i) whether the corruption applies to the {\em input} (i.e., action) or the {\em output} (i.e., reward function), (ii) whether {\em all samples} are corrupted, or only a {\em final reported point} is corrupted, and (iii) whether the corruptions are random or adversarial.  The case of random input noise on all samples was studied in \cite{beland2017bayes,nogueira2016unscented,dai2017stable}.  Perhaps closer to our work is \cite{martinez2018practical}, considering function outliers; however, no specific corruption model was adopted, and no theoretical regret bounds were given.

In \cite{bogunovic2018adversarially}, bounds on the simple regret are given for the case that the final reported input is adversarially perturbed, whereas the selected inputs are only subject to random output noise.  This makes it desirable to seek broad peaks, which bears some similarity to the input noise viewpoint \cite{beland2017bayes,nogueira2016unscented,dai2017stable} and level-set estimation \cite{gotovos2013active,bogunovic2016truncated}.  Our goal of attaining small {cumulative regret} under input perturbations requires very different techniques from these previous works.  Another distinct notion of robustness is considered in \cite{bogunovic2018robust}, in which some experiments in a batch may fail to produce an outcome.  None of the preceding works provide regret bounds in the case of non-stochastic corrupted observations.

%

\textbf{Related work on corrupted bandits.} Adversarially corrupted observations have recently been considered in the finite-arm stochastic MAB problem under various corruption models \cite{lykouris2018stochastic,gupta2019better,kapoor2019corruption}.  As mentioned above, \cite{lykouris2018stochastic} adopted a ``fast-slow'' algorithmic approach; this led to regret bounds of the form $R_T = O(KC \cdot R_{T}^{\textrm{non-c}} )$, where $R_{T}^{\textrm{non-c}}$ is a standard regret bound for the non-corrupted MAB setting.  In \cite{gupta2019better}, this bound was improved to $O(KC + R_{T}^{\textrm{non-c}})$ using an epoch-based approach in which the estimates of the arms' means are reset after each epoch, and the previous epoch guides which arms are selected in the next one.

Our algorithmic approach is based on that of \cite{lykouris2018stochastic}; however, the bulk of the theoretical analysis requires novel ideas.  In particular, our need to handle an infinite action space with correlated rewards between actions poses considerable challenges, as discussed above.  In more detail, we note the following:
\begin{itemize}
	\itemsep0em 
    \item Even when studying the case of a known corruption level (which is done as a stepping stone towards our main results), it is non-trivial to characterize the effect of the corruptions (see Lemma \ref{lemma:mean_difference} below);
    \item Characterizing that certain suboptimal points are never sampled after a certain time requires significant technical effort (see Lemmas \ref{lemma:F_queries_no_S_suboptimal_point} and \ref{lemma:time_to_suboptimality} below);
    \item We adopt a UCB-style approach (Alg. \ref{alg:universal_alg_given_C}) complementary to the elimination-style approach of \cite{lykouris2018stochastic}, and the former kind may be of independent interest even in the finite-arm setting.
\end{itemize}
In a parallel independent work \cite{li2019stochastic}, cumulative regret bounds were given for stochastic linear bandits, which are a special case of the GP setting (with a linear kernel).  The algorithm of \cite{li2019stochastic} is in fact more akin to that of \cite{gupta2019better}, which is potentially preferable due the latter attaining better bounds in the finite-arm setting.  However, the algorithm and results of \cite{li2019stochastic} crucially rely on the notion of {\em gaps} between the function values of corner points in the domain, and the idea of exploiting these gaps for linear bandits has no apparent generalization to the GP setting with general kernels.  In addition, even when we specialize to the linear kernel, neither our results nor those of \cite{li2019stochastic} imply each other, and the two both have benefits not provided by the other; see Appendix \ref{sec:comparison} for details.

\textbf{Outline.} We introduce the corruption-tolerant kernelized MAB problem in Section \ref{sec:problem}, and then present algorithms for three settings with increasing difficulty: Known corruption level (Section \ref{sec:known_C}), simultaneous handing of no corruption and a known corruption level (Section \ref{sec:known_or_zero}), and unknown corruption level (Section \ref{sec:unknown_C}).

\vspace*{-1ex}
\section{Problem Statement} \label{sec:problem}
\vspace*{-1ex}

We consider the problem of sequentially maximizing a fixed unknown function $f:D \rightarrow [-B_0, B_0]$, where $D \subset \R^d$ is a compact set and $B_0 > 0$. We assume that $D$ is endowed with a kernel function $k(\cdot, \cdot)$ defined on $D \times D$, and the kernel is normalized to satisfy $k(\vx, \vx') \leq 1$ for all $\vx,\vx' \in D$. We also assume that $f$ has a bounded norm in the corresponding Reproducing Kernel Hilbert Space (RKHS) $\Hc_k(D)$, i.e., $\|f \|_{k} \leq B$. This assumption permits the construction of confidence bounds via Gaussian process (GP) methods (see Lemma \ref{conf_lemma} below).

In the non-corrupted setting, at every time step $t$, we choose $\vx_t \in D$, and observe a noisy function value $y_t = f(\vx_t) + \epsilon_t$. In this work, we consider the corrupted setting, where we only observe an adversarially corrupted sample $\ty_t$. 
Formally, for each $t=1,\dots,T$:
\begin{itemize}[leftmargin=5ex,itemsep=0ex,topsep=0.25ex]
    \item Based on the previous decisions and corresponding corrupted observations $\lbrace (\vx_i, \ty_i) \rbrace_{i=1}^{t-1}$, the player selects a probability distribution $\Phi_t(\cdot)$ over $D$.
  \item Based on the knowledge of the true function $f$,\footnote{While knowing $f$ may appear to make the adversary overly strong, the defense mechanism in \cite{lykouris2018stochastic} for the finite-arm setting also implicitly allows the adversary to know the reward distributions.}  
  the previous decisions and corresponding observations $\lbrace (\vx_i, y_i) \rbrace_{i=1}^{t-1}$, and the player's distribution $\Phi_t(\cdot)$, the adversary chooses the corruptions $c_t(\cdot):D \rightarrow [-B_0, B_0]$.
  \item The agent draws $\vx_t \in D$ at random from $\Phi_t$, and observes the noisy and corrupted observation: 
  \begin{equation}
    \label{eq:corrupted_observation}
    \ty_t = y_t + c_t(\vx_t),
  \end{equation}
  where $y_t$ is the noisy non-corrupted observation: $y_t = f(\vx_t) + \epsilon_t$, where $\epsilon_t\sim \mathcal N(0,\sigma^2)$ with independence between times $t$.
\end{itemize}
Note that the adversary is allowed to be adaptive, i.e., the corruptions $c_t(\cdot)$ may depend on the agent's previously selected points and corresponding stochastic observations, as well as the distribution $\Phi_t(\cdot)$ of the player's next choice, but not its specific realization $\vx_t$.

We say that the problem instance is {\em $C$-corrupted} (i.e., the {\em corruption level} is $C$) if
\begin{equation}
\label{eq:total_corruption}
  \sum_{t=1}^T \max_{\vx \in D} |c_{t}(\vx)| \leq C.
\end{equation}
Clearly, when $C=0$, we recover the standard non-corrupted setting.
We measure the performance using the cumulative regret, which is also typically used in the non-corrupted bandit setting \cite{srinivas2009gaussian}:
\begin{equation}
  R_T = \sum_{t=1}^T \big( f(\vx^*) - f(\vx_t) \big), \label{eq:R_T}
\end{equation}
where $\vx^* = \argmax_{\vx \in D} f(\vx)$.  As noted in \cite{lykouris2018stochastic}, one could alternatively define the cumulative regret with respect to the corrupted values $\{ f(\vx) + c_t(\vx) \}$; the two notions coincide to within at most $2C$, and such a difference will be negligible in our regret bound anyway.  In Appendix \ref{sec:simple}, we outline how our results can be adapted for simple regret (i.e., the regret of a point reported at the end of $T$ rounds).

\vspace*{-1ex}
\subsection{Standard (non-corrupted) setting}
\vspace*{-1ex}

In the non-corrupted setting, existing algorithms use Gaussian likelihood models for the observations and zero-mean GP priors for modeling the uncertainty in $f$.
Posterior updates are performed according to a ``fictitious'' model in which the noise variables $\epsilon_t = y_t - f(\vx_t)$ are drawn independently across $t$ from $\mathcal{N}(0, \lambda)$, where $\lambda$ is a hyperparameter that may differ from the true noise variance $\sigma^2$. Given a sequence of inputs $\lbrace \vx_1, \dots, \vx_t \rbrace$ and their noisy observations $\lbrace y_1, \dots, y_t \rbrace$, the posterior distribution under this $\mathrm{GP}(\boldsymbol{0}, k)$ prior is also Gaussian, with the mean and variance
\begin{align}
\mu_{t}(\vx) &= \vk_t(\vx)^T\big(\vK_t + \lambda \mathbf{I}_t \big)^{-1} \vy_t,  \label{eq:posterior_mean} \\ 
\sigma_{t}^2(\vx) &= k(\vx,\vx) - \vk_t(\vx)^T \big(\vK_t + \lambda \mathbf{I}_t \big)^{-1} \vk_t(\vx), \label{eq:posterior_variance}
\end{align}
where $\vk_t(\vx) = \big[k(\vx_i,\vx)\big]_{i=1}^t$, and $\vK_t = \big[k(\vx_t,\vx_{t'})\big]_{t,t'}$ is the kernel matrix. Common kernels include the linear, squared exponential (SE) and Matérn kernels.

The main quantity that characterizes the regret bounds in the non-corrupted setting \cite{srinivas2009gaussian,chowdhury17kernelized} is the \emph{maximum information gain}, defined at time $t$ as 
\begin{equation}
\label{eq:max_info_gain}
  \gamma_t = \max_{\vx_1, \dots, \vx_t} \frac{1}{2} \ln  \det(\vI_t + \lambda^{-1}\vK_t).
\end{equation}
For compact and convex domains, $\gamma_t$ is sublinear in $t$ for various classes of kernels, e.g., $\mathcal{O}((\ln t)^{d+1})$ for the SE kernel, and $\mathcal{O}(t^{(d+1)d /((d+1)d + 2\nu)}\ln t))$ for the Mat\'ern kernel with $\nu > 1$ \cite{srinivas2009gaussian}.

The following well-known result of~\cite{abbasi2013online} provides confidence bounds around the unknown function in the non-corrupted setting.
\begin{lemma}
  \label{conf_lemma}
    Fix $f \in \Hc_k(D)$ with $\| f \|_k \leq B$, and consider the sampling model $y_t = f(\vx_t) + \epsilon_t$, with independent noise $\epsilon_t \sim \mathcal{N}(0,\sigma^2)$. Under the choice 
    \begin{equation} \label{eq:standard_beta}
      \beta_t = B + \sigma \lambda^{-1/2} \sqrt{2(\gamma_{t-1} + \ln(1 / \delta))}, 
    \end{equation}
    the following holds with probability at least $1 - \delta$:
    \begin{equation}    
        |\mu_{t-1}(\vx) - f(\vx)| \leq  \beta_t \sigma_{t-1}(\vx), \quad \forall \vx \in D, \forall t \geq 1, \label{eq:conf_bounds_std}
    \end{equation}
    where $\mu_{t-1}(\cdot)$ and $\sigma_{t-1}(\cdot)$ are given in \eqref{eq:posterior_mean} and \eqref{eq:posterior_variance}.
\end{lemma}
This lemma follows directly from \cite[Theorem~3.11]{abbasi2013online} (and \cite[Remark 3.13]{abbasi2013online}) and the definition \eqref{eq:max_info_gain} of $\gamma_t$. 

\begin{figure*}
          \centering
            \begin{subfigure}{.32\textwidth}
              \centering
              \includegraphics[scale=0.3]{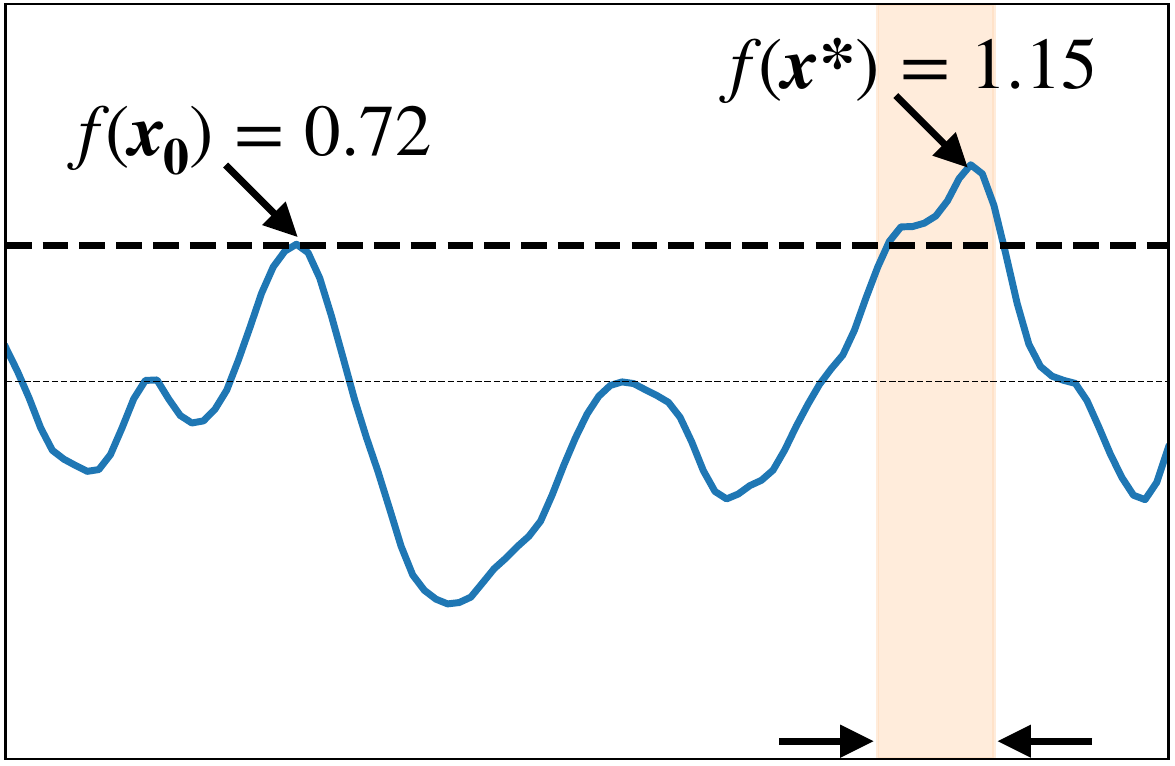}
            \end{subfigure}
            \begin{subfigure}{.32\textwidth}
              \centering
              \includegraphics[scale=0.3]{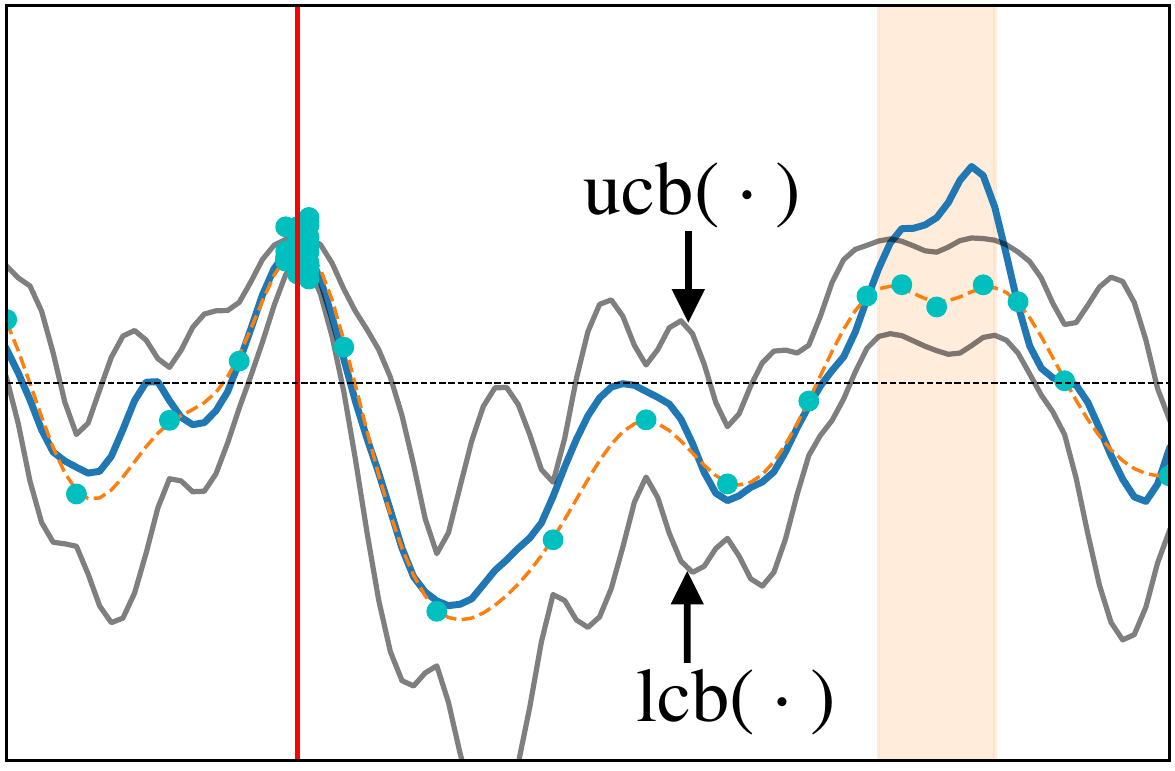}
            \end{subfigure}
            \begin{subfigure}{.32\textwidth}
              \centering
              \includegraphics[scale=0.3]{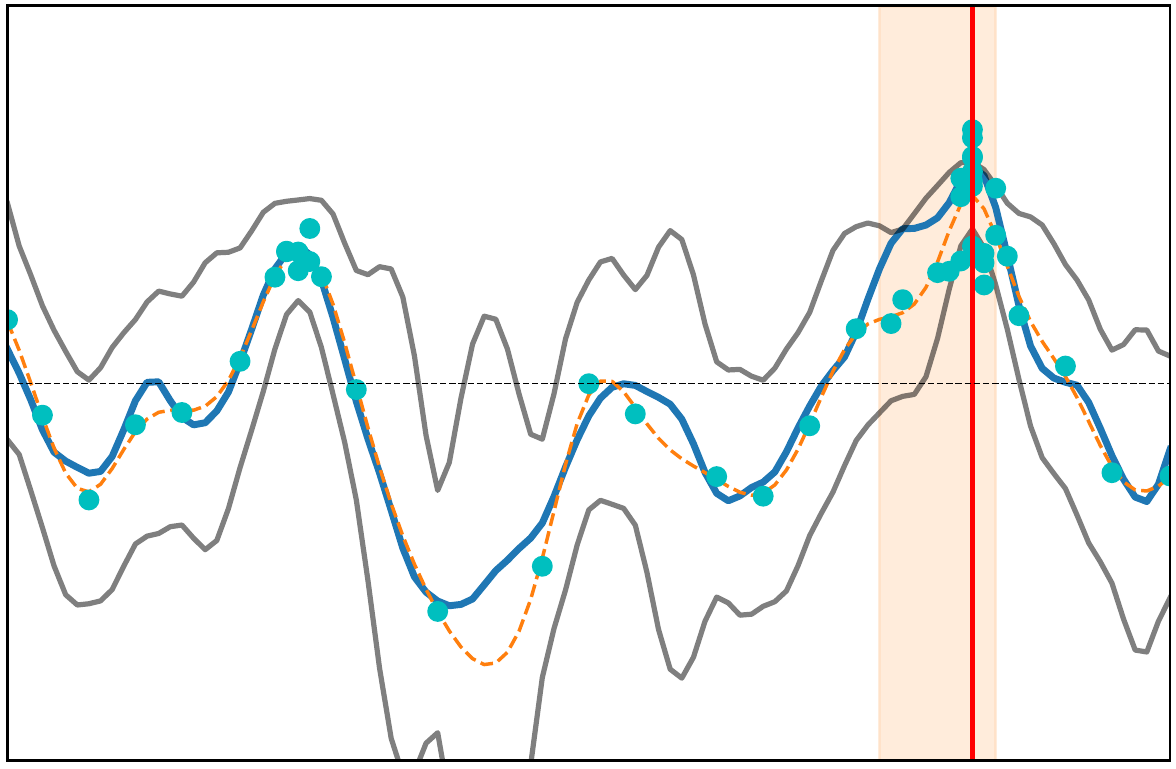}
            \end{subfigure}
            \caption{\small (Left) Function $f$, its global maximizer $\vx^*$, a local maximizer $\vx_0$, and the corruption region. (Middle) \gpucb~ eliminates the optimal region (and $\vx^*$) early on due to the corruptions, and continues sampling points in the suboptimal region around $\vx_0$. (Right) Our corruption-aware algorithm (see  Algorithm~\ref{alg:ucb_with_enlarged_conf_bounds}) does not eliminate the optimal region, and after the corruption budget is exhausted, it identifies the true maximizer $\vx^*$.}
            \vspace*{-1.5ex}
            \label{fig:failure_of_standard_methods}
\end{figure*}

\textbf{Lack of robustness against adversarial corruptions.} In the noisy non-corrupted setting, several algorithms have been developed and analyzed. A particularly well-known example is \gpucb, which selects
$\vx_t \in \argmax_{\vx \in D} \ucb_{t-1}(\vx):= \mu_{t-1}(\vx) + \beta_t \sigma_{t-1}(\vx)$.
\gpucb~achieves sublinear cumulative regret with high probability~\cite{srinivas2009gaussian,chowdhury17kernelized}, for a suitably chosen $\beta_t$ (e.g., as in~\eqref{eq:standard_beta}). Despite this success in the non-corrupted setting, these algorithms can fail under adversarial corruptions.

An illustrative example is provided in Figure~\ref{fig:failure_of_standard_methods}. Observations that correspond to the points sampled in the shaded region around the global maximizer $\vx^*$ are corrupted by the value $-f(\vx^*)/3$, up to a total corruption budget ($C=3.5$). In Figure~\ref{fig:failure_of_standard_methods} (Middle), the points selected by \gpucb~for $t = 50$ time steps are shown. \gpucb~eliminates the global maximizer early on due to corruptions, and later on, it only selects points from the suboptimal region and consequently suffers linear cumulative regret. In the subsequent sections, we design algorithms that are robust to corruptions, and are able to identify the true maximizer after the corruption budget $C$ is exhausted (see Figure~\ref{fig:failure_of_standard_methods} (Right)).

\vspace*{-1ex}
\section{Known Corruption Setting} \label{sec:known_C}
\vspace*{-1ex}

We first consider the case that the total corruption $C$ in~\eqref{eq:total_corruption} is known. Given a sequence of inputs $\lbrace \vx_1, \dots , \vx_t \rbrace$ and their corrupted observations $\lbrace \ty_1, \dots , \ty_t\rbrace$ (with $\ty_i = y_i + c_i(\vx_i)$), we form a posterior mean according to a $\mathrm{GP}(0, k(\vx, \vx'))$ prior and $\mathcal{N}(0,\lambda)$ sampling noise as follows:
\begin{align}
    \tilde{\mu}_{t}(\vx) &= \vk_t(\vx)^T(\vK_t + \lambda \vI)^{-1} \tilde{
    \vy}_t, \label{eq:corrupted_mean}
\end{align}
where $\tilde{\vy}_t = [\tilde{y}_1, \dots, \tilde{y}_t]$. Note that this matches the posterior mean formed in the non-corrupted setting, simply replacing $\vy_t$ by $\tilde{\vy}_t$.  In addition, we form the same posterior standard deviation $\sigma_{t-1}(\vx)$ as in the non-corrupted setting.  The role of the parameter $\lambda$ is discussed in Appendix \ref{sec:role_params}.

The following lemma provides an upper bound on the difference between the non-corrupted and corrupted posterior means, and is proved using the definitions of $\mu_t$ and $\tilde{\mu}_t$ along with RKHS function properties. All proofs can be found in the supplementary material.
\begin{lemma}
\label{lemma:mean_difference}
For any $\vx \in D$ and $t \geq 1$, we have
  $|\mu_{t-1} (\vx) - \tilde{\mu}_{t-1} (\vx)| \leq C \lambda^{-1/2}\sigma_{t-1}(\vx)$,
where $\mu_{t-1}(\cdot)$ and $\sigma_{t-1}(\cdot)$ are given in $\eqref{eq:posterior_mean}$ and $\eqref{eq:posterior_variance}$, and 
$\tilde{\mu}_{t-1} (\cdot)$ is given in~\eqref{eq:corrupted_mean}, with $\lambda > 0$.
\end{lemma}

By combining Lemmas~\ref{conf_lemma} and~\ref{lemma:mean_difference}, we obtain the following.

\begin{lemma}
  \label{conf_lemma_corrupted_known_C}
    Fix $f \in \Hc_k(D)$ with $\| f \|_k \leq B$. Under the choice $\beta_t^{(A_1)} = \beta_t + \lambda^{-1/2}C$ with $\beta_t$ given in \eqref{eq:standard_beta} and $\lambda > 0$, we have with probability at least $1 - \delta$ that
    \begin{equation}    
        |\tmu_{t-1}(\vx) - f(\vx)| \leq \beta_t^{(A_1)} \sigma_{t-1}(\vx), \quad \forall \vx \in D, \forall t \geq 1, \label{eq:conf_bounds}
    \end{equation}
    where $\tmu_{t-1}$ and $\sigma_{t-1}$ are given in \eqref{eq:corrupted_mean} and \eqref{eq:posterior_variance}.
\end{lemma}

In Algorithm~\ref{alg:ucb_with_enlarged_conf_bounds} ($A_1$), we present an upper confidence bound based algorithm with enlarged confidence bounds in accordance with Lemma \ref{conf_lemma_corrupted_known_C}.  We explicitly define these confidence bounds as follows:
\begin{align} \label{eq:lucbA1}
  \ucbalg(\vx) &= \tilde{\mu}_{t}(\vx) + \beta_{t+1}^{(A_1)} \sigma_{t}(\vx), \\
  \lcbalg(\vx) &= \tilde{\mu}_{t}(\vx) - \beta_{t+1}^{(A_1)} \sigma_{t}(\vx).
\end{align}
Once the validity of these confidence bounds is established via \eqref{eq:conf_bounds}, one can use standard analysis techniques \cite{srinivas2009gaussian} to bound the cumulative regret.  This is formally stated in the following.

\begin{lemma} \label{lem:conf_to_cumulative}
    Under the choice of $\beta_t^{(A_1)}$ in Lemma \ref{conf_lemma_corrupted_known_C} and $\lambda = 1$, conditioned on the event \eqref{eq:conf_bounds}, the cumulative regret incurred by Algorithm \ref{alg:ucb_with_enlarged_conf_bounds} satisfies $R_T = \mathcal{O}\big( \big(B + C + \sqrt{\ln(1/\delta)}\big) \sqrt{\gamma_T T} + \gamma_{T}\sqrt{T} \big)$.
\end{lemma}

\begin{algorithm}[!t]
    \caption{Gaussian Process UCB algorithm with known total corruption $C$} \label{alg:ucb_with_enlarged_conf_bounds}
    \begin{algorithmic}[1]
        \Require Prior $\mathrm{GP}(0,k)$, parameters $\sigma$, $\lambda$, $B$, $\lbrace \beta_t \rbrace_{t \ge 1}$, and total corruption $C$
        \For {$t = 1,2,\dotsc, T$}
        \State Set 
        \begin{equation} 
          \vx_t = \argmax_{\vx \in D}  \tilde{\mu}_{t-1} (\vx) + \beta_t^{(A_1)} \sigma_{t-1}(\vx),
        \label{eq:enlarged_ucb}
        \end{equation}
        $\quad$~~where $\beta_t^{(A_1)} = \beta_t + \lambda^{-1/2}C$
        \State Observe $\tilde{y}_t$ obtained via $\tilde{y}_t = f(\vx_t) + \epsilon_t + c_t(\vx_t)$  
        \State Update $\tilde{\mu}_t$ and $\sigma_t$ according to \eqref{eq:corrupted_mean} and \eqref{eq:posterior_variance} by including $(\vx_t,\tilde{y}_t)$ 
        \EndFor
        \State \textbf{end for}
    \end{algorithmic}
\end{algorithm}

The main theorem of this section is now obtained via a direct combination of Lemmas \ref{conf_lemma_corrupted_known_C} and \ref{lem:conf_to_cumulative}.

\begin{theorem}
\label{thm:known_C_thm}
    In the $C$-corrupted setting, Algorithm~\ref{alg:ucb_with_enlarged_conf_bounds} with $\lambda=1$ and $\beta_t^{(A_1)}$ set as in Lemma~\ref{conf_lemma_corrupted_known_C}, attains, with probability at least $1 - \delta$, cumulative regret
        $R_T = \mathcal{O}\big(\big(B + C + \sqrt{\ln(1 / \delta)}\big) \sqrt{\gamma_T T} + \gamma_{T}\sqrt{T} \big).$
\end{theorem}

Note that when $C=0$, this result recovers known non-corrupted cumulative regret bounds (\textit{cf.}~\cite[Theorem 3]{chowdhury17kernelized}). More generally, we can decompose the obtained regret bound into two terms: $R_T$ behaves as
\begin{equation} \label{eq:regret_decomposition}
  \mathcal{O}\Big(\underbrace{C\sqrt{\gamma_T T}}_{\text{due to corruption}} + \underbrace{ \big( B + \sqrt{\ln(1/\delta)} \big) \sqrt{\gamma_T T} + \gamma_{T}\sqrt{T}}_{\text{non-corrupted regret bound}}\Big).
\end{equation}
The obtained regret bound can be made more explicit by substituting the bound on $\gamma_T$ for particular kernels \cite{srinivas2009gaussian}, e.g., for the SE kernel we obtain $R_T = \mathcal{O}\big((C+B)\sqrt{T(\log T)^{d+1}} + (\log T)^{d+1} \sqrt{T}\big)$. 
In Appendix \ref{sec:dep_C}, we argue that the linear dependence on $C$ is unavoidable for any algorithm, and discuss cases where the dependence on $T$ is near-optimal.  However, we do not necessarily claim that the {\em joint} dependence on $C$ and $T$ is optimal; this is left for future work.

\vspace*{-1ex}
\section{Known-or-Zero Corruption Setting} \label{sec:known_or_zero}
\vspace*{-1ex}

In the previous section, we assumed that the upper bound $C$ on the total corruption is known and the problem is $C$-corrupted. In this section, we also assume that $C$ is known, but we consider a scenario in which the problem may be either $C$-corrupted or non-corrupted (i.e., the standard setting). Our goal is to develop an algorithm that has a similar guarantee to the previous section in the corrupted case, while also attaining a similar guarantee to \gpucb~\cite{srinivas2009gaussian} in the non-corrupted case, and thus obtaining strong guarantees in the two settings {\em simultaneously}.  Theorem \ref{thm:known_C_thm} fails to achieve this goal, since the regret depends on $C$ even if the problem is non-corrupted.   

\begin{algorithm}[!t]
\caption{\fsucb~algorithm}
\label{alg:universal_alg_given_C}
\begin{algorithmic}[1]
  \Require Prior $\mathrm{GP}(0,k)$, parameters $\sigma$, $\lambda$, $B$, $\alpha$, $\lbrace \beta_t^{(F)} \rbrace_{t \ge 1}$, $\lbrace \beta_t^{(S)} \rbrace_{t \ge 1}$, and total corruption $C$
    \State Initialize: $t_S, t_F := 1$, $\isValid = \text{True}$
    \For {$t = 1,2,\dotsc, T$}
      \If{$\isValid \textbf{ is } \text{True}$}
        \State Sample instance $A_t$: $A_t = S$ with probability $\min \lbrace 1, C^{-1}\rbrace$. Otherwise, $A_t = F$.
        \If{$A_t = F$}
        \State \hspace*{-2ex} $\vx_t \leftarrow \argmax_{\vx \in D}  \min_{A \in \lbrace F,S \rbrace} \ucbabar(\vx; 1)$ \label{line:acquisitionF}
        \Else
          \State \hspace*{-2ex}  $\vx_t \leftarrow \argmax_{\vx \in D} \ucbs(\vx; \alpha)$
        \EndIf
        \State Observe: $\ty_t = f(\vx_t) + c_t(\vx_t) + \epsilon_t$
        \State Set: $t_{A_{t}} \leftarrow t_{A_{t}} + 1$
        \State Update: $\tmu^{(A_t)}(\cdot)$, $\sigma^{(A_t)}(\cdot)$ to time $t_{A_{t}}$ by including $(\vx_t,\ty_t)$
        \If{$\min_{\vx} \big\lbrace \ucbftbar(\vx;1) - \lcbstbar(\vx;1) \big\rbrace < 0$}
          \State $\isValid \leftarrow \text{False}$ 
        \EndIf
      \Else
        \State Use all the collected data $\lbrace \vx_i, \ty_i \rbrace_{i=1}^{t}$ to compute $\tilde{\mu}_{t-1}(\cdot)$ and $\sigma_{t-1}(\cdot)$  
        \State Choose next point, observe and update according to Algorithm~\ref{alg:ucb_with_enlarged_conf_bounds}
      \EndIf
    \EndFor
\end{algorithmic}
\end{algorithm}

Our algorithm $\fsucb$ is described in Algorithm~\ref{alg:universal_alg_given_C}. It makes use of two instances labeled $F$ (fast; Line 6) and $S$ (slow; Line 8).
The $S$ instance is played with probability $1/C$, while the rest of the time $F$ is played. The intuition is that $F$ shrinks the confidence bounds faster but is not robust to corruptions, while $S$ is slower but robust to corruptions. We formalize this intuition below in Lemma \ref{lemma:corruption_observed_at_S} and \eqref{eq:confidence_alg2_1}--\eqref{eq:confidence_alg2_2}.

The instances use the following confidence bounds depending on an exploration parameter $\beta_{t_A+1}^{(A)}$ and an additional parameter $\alpha > 1$ whose role is discussed in Appendix \ref{sec:role_params} and after Lemma~\ref{lemma:time_to_suboptimality} below:
\begin{align}
  \ucbat(\vx; \alpha) &= \tilde{\mu}_{t_A}^{(A)}(\vx) + \alpha \beta_{t_A+1}^{(A)} \sigma_{t_A}^{(A)}(\vx) \label{eq:ucbA} \\
  \lcbat(\vx; \alpha) &= \tilde{\mu}_{t_A}^{(A)}(\vx) - \alpha \beta_{t_A+1}^{(A)} \sigma_{t_A}^{(A)}(\vx), \label{eq:lcbA}
\end{align}
where $t_A$ is the number of times an instance $A \in \{F,S\}$ has been selected at a given time instant.  We also make use of the following {\em intersected} confidence bounds, which have the convenient feature of being monotone:
\begin{align} \label{eq:lucbAbar}
  \ucbatbar(\vx; \alpha) &= \min_{t'_A \le t_A} \ucbap(\vx; \alpha), \\
  \lcbatbar(\vx; \alpha) &= \max_{t'_A \le t_A} \lcbap(\vx; \alpha).
\end{align}

In \fsucb, we check if the following condition (Line 12) holds:
\begin{equation}
\label{eq:condition}
  \min_{\vx \in D} \left\lbrace \ucbftbar(\vx;1) - \lcbstbar(\vx;1) \right\rbrace < 0. 
\end{equation}

In the non-corrupted setting, under the high-probability event in Lemma \ref{conf_lemma} (for both $F$ and $S$), this condition never holds.  Hence, when it does hold, we have detected that the problem is $C$-corrupted.
In such a case, the algorithm permanently switches to running Algorithm~\ref{alg:ucb_with_enlarged_conf_bounds} with $C$ as the input. Note that we can check the condition in \eqref{eq:condition} by using a global optimizer to find a minimizer of $g(\vx):= \ucbftbar(\vx;1) - \lcbstbar(\vx;1)$, and checking whether its value is smaller than $0$.

Finally, the inner minimization over $A \in \lbrace F,S \rbrace$ in the $F$ instance, together with the validity of the condition~\eqref{eq:condition}, ensures that $F$ does not select a point that is already ``ruled out'' by the robust instance $S$. We make this statement precise in Lemma~\ref{lemma:F_queries_no_S_suboptimal_point} below.  

\vspace*{-1ex}
\subsection{Analysis}
\vspace*{-1ex}

First, we provide a high-probability bound on the total corruption that is observed by the $S$ instance. Specifically, we show that because it is sampled with probability $1/C$, the total corruption observed by $S$ is constant with high probability, i.e.,~it is upper bounded by a value not depending on $T$. 
\begin{lemma} \label{lemma:corruption_observed_at_S}
    The $S$ instance in \fsucb~observes, with probability at least $1 - \delta$, a total corruption $\sum_{t=1}^T |c_t(\vx_t)| \bone\{ A_t = S \}$ of at most $3 + B_0\ln (1/\delta)$.
\end{lemma}

We now fix a constant $\delta \in (0,1)$ and condition on three high-probability events:
\begin{enumerate}[leftmargin=5ex,itemsep=0ex,topsep=0.25ex]
  \item If $\beta^{(F)}_{t_F} = B + \sigma \lambda^{-1/2} \sqrt{2\left(\gamma_{t_F - 1} + \ln \left( \frac{5}{\delta} \right)\right)}$ and the setting is non-corrupted, the following holds with probability at least $1 - \frac{\delta}{5}$:
    \begin{equation} \label{eq:confidence_alg2_1}
      \lcbf(\vx; 1) \leq f(\vx) \leq \ucbf(\vx; 1),
    \end{equation}
  for all $\vx \in D$ and $t_F \geq 1$.
  This claim follows from Lemma~\ref{conf_lemma} by setting the corresponding failure probability to~$\frac{\delta}{5}$.
  
  \item If $\beta^{(S)}_{t_S} = B + \sigma \lambda^{-1/2} \sqrt{2\left(\gamma_{t_S - 1} + \ln \left( \frac{5}{\delta} \right)\right)} + \lambda^{-1/2} (3 + B_0 \ln \left( \frac{5}{\delta} \right))$, then the following holds in both the non-corrupted and corrupted settings 
    with probability at least $1 - \frac{2 \delta}{5}$:
    \begin{equation} \label{eq:confidence_alg2_2}
      \lcbs(\vx; 1) \leq f(\vx) \leq \ucbs(\vx; 1),
    \end{equation} 
  for all $\vx \in D$ and $t_S \geq 1$.
  This follows from Lemmas~\ref{conf_lemma_corrupted_known_C} and~\ref{lemma:corruption_observed_at_S} (using $3 + B_0 \ln \left( \frac{5}{\delta}\right)$ in place of $C$ in Lemma \ref{conf_lemma_corrupted_known_C}), by setting the corresponding failure probabilities to $\frac{\delta}{5}$ in both. Taking the union bound over these two events establishes the claim. Note that \eqref{eq:confidence_alg2_2} corresponds to $\alpha = 1$, but directly implies an analogous condition for all $\alpha>1$ (since increasing $\alpha$ widens the confidence region \eqref{eq:ucbA}--\eqref{eq:lcbA}).
  \item If the condition in~\eqref{eq:condition} is detected at any time instant, then Algorithm~\ref{alg:universal_alg_given_C} permanently switches to running Algorithm~\ref{alg:ucb_with_enlarged_conf_bounds}. If Algorithm~\ref{alg:ucb_with_enlarged_conf_bounds} is run with $\beta_t^{(A_1)} = B + \sigma \lambda^{-1/2} \sqrt{2\left(\gamma_{t - 1} + \ln \left( \frac{5}{\delta} \right)\right)} + \lambda^{-1/2}C$, then with probability at least $1 - \frac{\delta}{5}$: 
    \begin{equation} \label{eq:confidence_alg2_3}
      \lcb_{t- 1}^{(A_1)}(\vx) \leq f(\vx) \leq \ucb^{(A_1)}_{t- 1}(\vx),
    \end{equation}
    for all $\vx \in D$ and $t \geq 1$, under the definitions in \eqref{eq:lucbA1}.
  This is by Lemma~\ref{conf_lemma_corrupted_known_C} with $\frac{\delta}{5}$ in place of $\delta$. 
\end{enumerate}
By the union bound, \eqref{eq:confidence_alg2_1}--\eqref{eq:confidence_alg2_3} all hold with probability at least $1 - \frac{4\delta}{5}$.   In addition, by the definitions in \eqref{eq:lucbAbar}, these properties remain true when $\ucb^{(A)}$ and $\lcb^{(A)}$ are replaced by $\ucbbar^{(A)}$ and $\lcbbar^{(A)}$.

The confidence bounds of $F$ are only valid in the non-corrupted case, and hence, in the case of corruptions we rely on the confidence bounds of $S$.
Specifically, we show that $F$ never queries a point that is strictly suboptimal according to the confidence bounds of $S$.
\begin{lemma} \label{lemma:F_queries_no_S_suboptimal_point}
  Suppose that~\eqref{eq:confidence_alg2_1} and~\eqref{eq:confidence_alg2_2} hold. For any time $t \geq 1$, if $A_t = F$ in \fsucb, then the selected point $\vx_t \notin \Sc_{t_S}$, where 
  \begin{multline}
    \Sc_{t_S} = \lbrace \vx \in D:\; \exists \vx' \in D,\; \\ \lcbsbar(\vx'; 1) > \ucbsbar(\vx; 1) \rbrace \label{eq:def_St}
  \end{multline}
  represents the set of strictly suboptimal points according to the intersected S-confidence bounds.
\end{lemma}

By the monotonicity of $\lcbsbar$ and $\ucbsbar$, the set $\Sc_{t_S}$ is non-shrinking in $t$. The proof shows that $F$ always favors $\vx'$ from \eqref{eq:def_St} over $\vx \in \Sc_{t_S}$, i.e., $\vx'$ has a higher value of $\min_{A \in \lbrace F,S \rbrace} \ucbabar(\cdot;1)$  
(see Line \ref{line:acquisitionF} of Algorithm \ref{alg:universal_alg_given_C}).  To show this, we upper bound $\min_{A \in \lbrace F,S \rbrace} \ucbabar(\vx;1)$ in terms of $\lcbsbar$ via \eqref{eq:def_St}, and lower bound $\min_{A \in \lbrace F,S \rbrace} \ucbabar(\vx;1)$ in terms of $\lcbsbar$ via the confidence bounds and condition \eqref{eq:condition}.

The next lemma characterizes the number of queries made by the $S$ instance before a suboptimal point becomes ``eliminated'', i.e., the time after which the point belongs to $\Sc_{t_S}$.

\begin{lemma}\label{lemma:time_to_suboptimality}
    Suppose that the $S$ instance is run with $\beta^{(S)}_{t_S}$ corresponding to~\eqref{eq:confidence_alg2_2} and $\alpha=2$.  Then, conditioned on the high-probability confidence bounds in \eqref{eq:confidence_alg2_2}, for any given suboptimal point $\vx \in D$ such that $f(\vx^*) - f(\vx) \geq \Delta_0 > 0$, it holds that $\vx \in \Sc_{t_S}$ after 
    \begin{equation} \label{eq:time_S}
      t_S = \min\bigg\{ \tau \,:\, \sqrt{ \tfrac{ 16 \alpha^2 (\beta^{(S)}_{\tau})^{2} \gamma_{\tau}}{\tau}} \le \tfrac{\Delta_0}{10} \bigg\}.
    \end{equation}
\end{lemma}

This lemma's proof is perhaps the trickiest, and crucially relies on the fact that $\alpha > 1$. We show that by the time given in \eqref{eq:time_S}, we have encountered a round $i$ in which a $\frac{\Delta_0}{10}$-optimal point $\vx_i$ is queried with the confidence width also being at most $\frac{\Delta_0}{10}$.  This means that $\vx_i$ is much closer to optimal than the $\Delta_0$-suboptimal point $\vx$ in the lemma statement.  Using the fact that $\vx_i$ had a higher UCB score than $\vx$, we can also deduce that the posterior standard deviation at $\vx$ was not too large.  Since replacing $\alpha = 2$ by $\alpha = 1$ (as done in the definition of $\Sc_{t_S}$ in \eqref{eq:def_St}) halves the confidence width, we can combine the above findings to deduce that the confidence bounds indeed rule out $\vx$ at time $i < t_S$, and hence also for all subsequent times due to the monotonicity of the confidence bounds.

Finally, we state the main theorem of this section, whose proof combines the preceding lemmas.

  \begin{theorem} \label{thm:corrupted_vs_stochastic_thm}
    For any $f \in \Hc_k(D)$ with $\|f \|_{k} \leq B$, let $\delta \in (0,1)$, and consider \fsucb~run with $\alpha = 2$, $\lambda=1$,
    \begin{align}
      \beta^{(F)}_{t_F} &= B + \sigma \sqrt{2\left(\gamma_{t_F - 1} + \ln \left( \tfrac{5}{\delta}   \right)\right)}, \label{eq:beta_F} \\
      \beta^{(S)}_{t_S} &= B + \sigma \sqrt{2\left(\gamma_{t_S - 1} + \ln \left( \tfrac{5}{\delta} \right)\right)} + (3 + B_0 \ln \left( \tfrac{5}{\delta} \right)), \label{eq:beta_S}
    \end{align}
    and $\beta_t^{(A_1)}$ set in Algorithm~\ref{alg:ucb_with_enlarged_conf_bounds} as 
    $\beta_t^{(A_1)} = B + \sigma \sqrt{2\left(\gamma_{t - 1} + \ln \left( \tfrac{5}{\delta} \right)\right)} + C$.
    Then, after $T$ rounds, with probability at least $1-\delta$ the cumulative regret satisfies
    \begin{equation}
        R_T = \mathcal{O}\left(\Big(B + B_0\ln (\tfrac{1}{\delta}) + \sqrt{\ln (\tfrac{1}{\delta})}\Big) \sqrt{T\gamma_T} + \gamma_T \sqrt{T}\right)
    \end{equation}
    in the non-corrupted case and 
    \begin{multline}
        R_T =\mathcal{O}\bigg( (1+C)\ln(\tfrac{T}{\delta})\Big( \Big(B + B_0\ln (\tfrac{1}{\delta}) + \sqrt{\ln (\tfrac{1}{\delta})}\Big) \\ \times \sqrt{\gamma_{T}T} + \gamma_{T} \sqrt{T} \Big) \bigg) \label{eq:RT_corr_final}
    \end{multline}
    in the corrupted case.
  \end{theorem}

The non-corrupted case is straightforward to prove, essentially applying standard arguments separately to $F$ and $S$. The corrupted case requires more effort.  Lemma \ref{lemma:time_to_suboptimality} characterizes the time after which points with a given regret are no longer sampled by $S$, which permits bounding the cumulative regret incurred by $S$.  By Lemma \ref{lemma:F_queries_no_S_suboptimal_point}, the points in $\Sc_{t_S}$ are also not sampled by $F$, and on average $F$ is played at most $C$ times more frequently than $S$.  Converting this average to a high-probability bound using basic concentration, this factor of $C$ becomes $C\ln(\tfrac{5T}{\delta})$, and we obtain \eqref{eq:RT_corr_final}.

Using the notation $\tilde{\mathcal{O}}(\cdot)$ to hide logarithmic factors, the bound obtained in the non-corrupted case simplifies to $R_T = \tilde{\mathcal{O}}\big((B + B_0) \sqrt{T\gamma_T} + \gamma_T \sqrt{T}\big)$, and unlike the result from Theorem~\ref{thm:known_C_thm}, it does not depend on $C$. The obtained bound is only a constant factor away from the standard non-corrupted one (\emph{cf.}~\eqref{eq:regret_decomposition}), while at the same time our algorithm achieves $R_T = \tilde{\mathcal{O}}\big( C(B + B_0) \sqrt{T\gamma_T} + C\gamma_T \sqrt{T}\big)$ in the $C$-corrupted case. 
As before, we can make the results obtained in this theorem more explicit by substituting the bounds for $\gamma_T$ for various kernels of interest \cite{srinivas2009gaussian}.


\vspace*{-1ex}
\section{Unknown Corruption Setting} \label{sec:unknown_C}
\vspace*{-1ex}

In this section, we assume that the total corruption $C$ defined in~\eqref{eq:total_corruption} is unknown to the algorithm. 
Despite this additional challenge, most of the details are similar to the known-or-zero setting, so to avoid repetition, we omit some details and focus on the key differences.

\textbf{Algorithm.} Our corruption-agnostic algorithm is shown in Algorithm~\ref{alg:unknown_C_alg}.  We again take inspiration from the finite-arm counterpart \cite{lykouris2018stochastic}, considering {\em layers} $\ell=1,\dotsc,\lceil \log_2 T \rceil$ that are sampled with probability $2^{-\ell}$ (with any remaining probability going to layer $1$).  The idea is that any layer with $2^{\ell} \ge C$ is robust, for the same reason that the $S$ instance is robust in $\fsucb$ (Algorithm \ref{alg:universal_alg_given_C}).

Each instance $\ell$ makes use of confidence bounds defined as follows for some parameters $\beta_{t_{\ell}}^{(\ell)}$ to be chosen later:
\begin{align}
  \ucb^{(\ell)}_{t_{\ell}}(\vx; \alpha) &= \tilde{\mu}_{t_\ell}(\vx) + \alpha \beta_{t_{\ell}+1}^{(\ell)} \sigma_{t_{\ell}}(\vx) \\
  \lcb^{(\ell)}_{t_{\ell}}(\vx; \alpha) &= \tilde{\mu}_{t_\ell}(\vx) - \alpha \beta_{t_{\ell}+1}^{(\ell)} \sigma_{t_\ell}(\vx),
\end{align}
where $t_{\ell}$ denotes the number of times instance $\ell$ has been selected by time $t$, and $\alpha > 1$. Similarly to the Section \ref{sec:known_or_zero}, we define intersected confidence bounds:
\begin{align}
	\ucbltbar(\vx; \alpha) &= \min_{t'_\ell \le t_\ell} \ucblp(\vx; \alpha) \\
	\lcbltbar(\vx; \alpha) &= \max_{t'_\ell \le t_\ell} \lcblp(\vx; \alpha).
\end{align}
Each instance $\ell$ selects a point according to $\argmax_{\vx \in M_t^{(\ell)}} \ucb^{(\ell)}_{t_{\ell}-1}(\vx; \alpha)$, where $M_t^{(\ell)}$ represents a set of {\em potential maximizers} at time $t$, i.e., a set of points that could still be the global maximizer according to the confidence bounds. More formally, these sets are defined recursively as follows:\footnote{Note that a given set $M_t^{(\ell)}$ may be non-convex, making the constraint $\vx \in D$ in the UCB rule non-trivial to enforce in practice (e.g., one may use a discretization argument).  Our focus is on the theory, in which we assume that the acquisition function can be optimized exactly.}
\begin{align}
	\hspace*{-3ex} &M_t^{(\ell)} :=  \Big\lbrace \vx \in D: \ucbbar^{(\ell)}_{t_{\ell} - 1}(\vx;1) \geq \max_{\vx' \in D} \lcbbar^{(\ell)}_{t_{\ell}-1}(\vx';1) \Big\rbrace \nonumber \\
         &\hspace*{4cm} \text{for } \ell = \lceil \log_2 T \rceil,\\
	\hspace*{-3ex} &M_t^{(\ell)} := M_{t}^{(\ell + 1)} \cap \Big\lbrace \vx \in D: \nonumber \\
        &\ucbbar^{(\ell)}_{t_{\ell} - 1}(\vx;1) \geq \max_{\vx' \in D} \lcbbar^{(\ell)}_{t_{\ell}-1}(\vx';1) \Big\rbrace  ~~ \text{for }\ell < \lceil \log_2 T \rceil.
\end{align}
Two key properties of these sets are: (i) $M^{(\ell)}_{t} \subseteq M^{(\ell)}_{t'}$ for every $t > t'$ and $\ell \in \lbrace 1, \dots, \lceil \log_2 T \rceil \rbrace$ due to the monotonicity of the confidence bounds; and (ii) $M_t^{(1)} \subseteq M_t^{(2)} \dots \subseteq M_t^{(\lceil \log_2 T \rceil)}$ for every $t$. The latter property implies that once a point is eliminated at some layer $\ell$, 
it is also eliminated from all $M_t^{(1)}, \dots, M_t^{(\ell-1)}$, while the former property ensures that it remains eliminated for all subsequent time steps $\lbrace t+1, \dots, T \rbrace$.

Similarly to $\fsucb$, each layer uses $\ucb^{(\ell)}_{t_{\ell}-1}(\vx; \alpha)$ {\em with $\alpha$ strictly larger than $1$} (e.g., $\alpha = 2$ suffices) in its acquisition function, while replacing $\alpha$ by $1$ in the confidence bounds when constructing the set of potential maximizers.  This is done to permit the application of Lemma \ref{lemma:time_to_suboptimality}; the intuition behind doing so is discussed in Appendix \ref{sec:role_params} and following Lemma \ref{lemma:time_to_suboptimality}.

In the case that $M_t^{(\ell)}$ corresponding to the selected $\ell$ at time $t$ is empty, the algorithm finds the lowest layer $i$ for which $M^{(i)} \neq \emptyset$, and selects the point that maximizes that layer's upper confidence bound. In this case, the algorithm makes no changes to the confidence bounds or the sets of potential maximizers. 

\begin{algorithm}[t!]
\caption{\fsucb~algorithm with Unknown Corruption Level $C$}
\label{alg:unknown_C_alg}
\begin{algorithmic}
	\Require Prior $\mathrm{GP}(0,k)$, parameters $\sigma$, $\lambda$, $B$, $\alpha$, $\lbrace \beta_{t_{\ell}}^{(\ell)} \rbrace_{t \ge 1}$ for all $\ell \in \lbrace 1, \dots, \lceil \log_2 T \rceil \rbrace$
    \State Initialize: For all $\ell \in \lbrace 1, \dots, \lceil \log_2 T \rceil \rbrace$, set $M_1^{(\ell)} = D$ 
    \For {$t = 1,2,\dotsc, T$}
    	\State Sample instance $\ell \in \lbrace 1, \dots, \lceil \log_2 T \rceil \rbrace$ w.p. $2^{-\ell}$. With remaining prob., sample $\ell=1$.
    			\If{$M_t^{(\ell)} \neq \emptyset$}
		    		\State $\vx_t \leftarrow \argmax_{\vx \in M_t^{(\ell)}} \ucb^{(\ell)}_{t_{\ell}-1}(\vx; \alpha)$
		    		\State Observe $\ty_t = f(\vx_t) + c_t(\vx_t) + \epsilon_t$
	    			\State Update $\tmu^{(\ell)}(\cdot)$, $\sigma^{(\ell)}(\cdot)$ by including $(\vx_t, \ty_t)$
	    			\State $t_{\ell} \leftarrow t_{\ell} + 1$
	    			\State $M_{t+1}^{(\ell)} \leftarrow \lbrace \vx \in D: \ucbbar^{(\ell)}_{t_{\ell} - 1}(\vx;1) \geq \max_{\vx' \in D} \lcbbar^{(\ell)}_{t_{\ell}-1}(\vx';1) \rbrace$
	    			\State $M_{t+1}^{(i)} \leftarrow M^{(\ell)}_{t+1} \cap M^{(i)}_t$ for $i \in \lbrace 1, \dots, \ell-1 \rbrace$
	    			\State $M_{t+1}^{(i)} \leftarrow M^{(i)}_{t}$ for $i \in \lbrace \ell+1, \dots, \lceil \log T \rceil \rbrace$
	    		\Else
	    			\State $\ell \leftarrow \argmin_{i \in \lbrace \ell+1, \dots, \lceil \log_2 T \rceil \rbrace} \lbrace M_t^{(i)} \neq \emptyset \rbrace$
	    			\State $\vx_t \leftarrow \argmax_{\vx \in M_t^{(\ell)}} \ucb^{(\ell)}_{t_{\ell}-1}(\vx; \alpha)$
		    		\State Observe: $\ty_t = f(\vx_t) + c_t(\vx_t) + \epsilon_t$
		    		\State $M_{t+1}^{(i)} \leftarrow M^{(i)}_{t}$ for every $i \in \lbrace 1, \dots, \lceil \log_2 T \rceil \rbrace$
	    		\EndIf
    \EndFor
\end{algorithmic}
\end{algorithm}

\textbf{Regret bound.}  With $\fsucb$ and its theoretical analysis in place, we can also obtain a near-identical regret bound in the case of unknown $C$.  We only provide a brief outline here, with further details in the supplementary material.


We let the robust layer $\ell^* = \lceil \log_2 C \rceil$ play the role of $F$ and eliminate suboptimal points.  Since $2^{-\ell^*} \ge \frac{1}{2C}$, the regret incurred in the lower layers is at most a factor $2C$ higher than that of layer $\ell^*$ on average, and this leads to a similar analysis to that used in the proof of Theorem \ref{thm:corrupted_vs_stochastic_thm}.  Our final main result is stated as follows.

\begin{theorem} \label{thm:unkownC}
    For any $f \in \Hc_k(D)$ with $\|f \|_{k} \leq B$, and any $\delta \in (0,1)$, under the parameters
    \begin{multline}
        \beta^{(\ell)}_{t_{\ell}} = B + \sigma \sqrt{2\left(\gamma_{t_\ell - 1} + \ln \left( \frac{4 (1 + \log_2 T)}{\delta} \right)\right)} \\ + 3 + B_0 \ln \left( \frac{4(1 + \log_2 T)}{\delta} \right),
    \end{multline}
    we have that for any unknown corruption level $C > 0$, the cumulative regret of Algorithm \ref{alg:unknown_C_alg} satisfies
    \begin{multline}
    	R_T = \mathcal{O}\bigg( (1+C)\ln(\tfrac{T}{\delta}) \\ \times\Big( \Big(B + B_0\ln (\tfrac{\log T}{\delta}) + \sqrt{\ln (\tfrac{\log T}{\delta})}\Big) \sqrt{\gamma_{T}T} + \gamma_{T} \sqrt{T} \Big) \bigg)
    \end{multline}
    with probability at least $1-\delta$.
\end{theorem}

This has the same form as \eqref{eq:RT_corr_final}, with $\frac{\delta}{\log T}$ in place of $\delta$ (since there are $\lceil \log_2 T\rceil$ layers). 

\vspace*{-1ex}
\section{Conclusion}
\vspace*{-1ex}

We have introduced the kernelized MAB problem with adversarially corrupted samples.  We provided novel algorithms based on enlarged confidence bounds and randomly-selected fast/slow instances that are provably robust against such corruptions, with the regret bounds being linear in the corruption level.  To our knowledge, we are the first to handle this form of adversarial corruption in any bandit problem with an infinite action space and correlated rewards, which are two key notions that significantly complicate the analysis.

An immediate direction for further research is to better understand the {\em joint} dependence on the corruption level $C$ and time horizon $T$.  The linear $O(C)$ dependence is unavoidable (see Appendix \ref{sec:dep_C}), and the $O( B\sqrt{\gamma_T T} + \gamma_T \sqrt{T} )$ dependence matches well-known bounds for the non-corrupted setting \cite{srinivas2009gaussian,chowdhury17kernelized} (in some cases having near-matching lower bounds \cite{scarlett2017lower}), but it is unclear whether the \emph{product} of these two terms is unavoidable.  

\subsubsection*{Acknowledgments}
This work was gratefully supported by Swiss National Science Foundation, under the grant SNSF NRP 75 407540\_167189, ERC grant 815943 and ETH Z\"urich Postdoctoral Fellowship 19-2 FEL-47.
J. Scarlett was supported by the Singapore National Research Foundation (NRF) under grant number R-252-000-A74-281.

\bibliography{mybib}

\onecolumn
\newpage
\appendix

{\centering
    {\huge \bf Supplementary Material}
    
    {\Large \bf  Corruption-Tolerant Gaussian Process Bandit Optimization \\ [2mm] {\normalsize \bf {Ilija Bogunovic, Andreas Krause, Jonathan Scarlett (AISTATS 2020)} \par }  
}}

All citations below are to the reference list in the main document.

\section{Proof of Lemma~\ref{lemma:mean_difference} (Corrupted vs.~Non-Corrupted Posterior Mean)}

Our analysis uses techniques from \cite[Appendix~C]{chowdhury17kernelized}.
Let $\vx$ be any point in $D$, and fix a time index $t \geq 1$. From the definitions of $\tilde{\mu}_{t}(\cdot), \mu_{t}(\cdot)$ and $\ty_t$ (Eq.~\eqref{eq:posterior_mean},~\eqref{eq:corrupted_mean} and~\eqref{eq:corrupted_observation}), we have
\begin{align}
		\tilde{\mu}_{t}(\vx) &= \vk_t(\vx)^T(\vK_t + \lambda \vI_t)^{-1} \tilde{
		\vy}_t \\
		&=  \vk_t(\vx)^T(\vK_t + \lambda \vI_t)^{-1} \vy_t +  \vk_t(\vx)^T(\vK_t + \lambda \vI_t)^{-1} \vc_t\\
		&=  \mu_{t}(\vx) + \vk_t(\vx)^T(\vK_t + \lambda \vI_t)^{-1} \vc_t, \label{eq:term_of_interest}
\end{align}
where $\tilde{\vy}_t = [\tilde{y}_1, \dots, \tilde{y}_t]^T$ and $\vc_t = [c_1(\vx_1), \dots, c_t(\vx_t)]^T$.
We proceed by upper bounding the absolute difference between $\tilde{\mu}_{t}(\vx)$ and $\mu_{t}(\vx)$, i.e,
$|\vk_t(\vx)^T(\vK_t + \lambda \vI_t)^{-1} \vc_t|$.

Let $\Hc_{k}(D)$ denote the RKHS associated with the kernel $k$ and domain $D$. 
We define $\phi(\vx) := k(\vx,\cdot)$, where $\phi \,:\, D \rightarrow \Hc_{k}(D)$ maps $\vx \in D$ to the RKHS associated with the kernel. For any two functions $f_1, f_2$ in $\Hc_{k}(D)$, we write $f_1^T f_2$ to denote the kernel inner product $\langle f_1, f_2 \rangle_k$, which implies that $\|f\|_k = \sqrt{f^Tf}$. By the RKHS reproducing property, i.e., $f(\vx) = \langle f, k(\vx, \cdot) \rangle_k$ for all $\vx \in D$, and the fact that $k(\vx,\cdot) \in \Hc_{k}(D)$ for all $\vx \in D$, we can write
\[
	k(\vx, \vx') = \langle k(\vx,\cdot), k(\vx',\cdot) \rangle_k = \langle \phi(\vx), \phi(\vx') \rangle_k = \phi(\vx)^T\phi(\vx')
\]
for all $\vx, \vx' \in D$. It also follows that $\vK_t = \Phi_t \Phi_t^T$ where $\Phi_t = [\phi(\vx_1), \dots, \phi(\vx_t)]^T$, and $k_t(\vx) = \Phi_t \phi(\vx)$.  (Here and subsequently, the notation $f_1^T f_2 = \langle f_1, f_2 \rangle_k$ similarly extends to matrix multiplication operations.)

Using these properties, we can characterize the second term of \eqref{eq:term_of_interest} as follows:
\begin{align}
	&|\vk_t(\vx)^T(\vK_t + \lambda \vI_t)^{-1} \vc_t| \nonumber \\
    &\qquad = |\phi(\vx)^T \Phi_t^T(\Phi_t \Phi_t^T + \lambda \vI_t)^{-1} \vc_t| \label{eq:bigstep1} \\
	&\qquad = |\langle \phi(\vx)^T (\Phi_t^T \Phi_t + \lambda \vI_{\Hc_k})^{-1}, \Phi_t^T \vc_t\rangle_k| \label{eq:bigstep2} \\
	&\qquad\leq \|(\Phi_t^T \Phi_t + \lambda \vI_{\Hc_k})^{-1/2}\phi(\vx)\|_k \|(\Phi_t^T \Phi_t + \lambda \vI_{\Hc_k})^{-1/2}\Phi_t^T \vc_t\|_k \label{eq:bigstep3} \\
	&\qquad = \sqrt{\phi(\vx)^T(\Phi_t^T \Phi_t + \lambda \vI_{\Hc_k})^{-1}\phi(\vx)} \sqrt{(\Phi_t^T \vc_t)^T(\Phi_t^T \Phi_t + \lambda \vI_{\Hc_k})^{-1}\Phi_t^T \vc_t} \label{eq:bigstep4} \\
	&\qquad= \lambda^{-1/2}\sigma_{t}(\vx) \sqrt{\vc_t^T \Phi_t\Phi_t^T(\Phi_t \Phi_t^T + \lambda \vI_t)^{-1} \vc_t} \label{eq:variance_enters} \\
	&\qquad =  \lambda^{-1/2}\sigma_{t}(\vx) \sqrt{\vc_t^T \vK_t(\vK_t + \lambda \vI_t)^{-1} \vc_t} \label{eq:bigstep6} \\
	&\qquad \leq  \lambda^{-1/2}\sigma_{t}(\vx) \sqrt{\lambda_{\text{max}}\left(\vK_t(\vK_t + \lambda \vI_t)^{-1}\right) \|\vc_t \|^2_2} \label{eq:max_ev_enters} \\
	&\qquad \leq  \lambda^{-1/2}\sigma_{t}(\vx) C \sqrt{\lambda_{\text{max}}\left(\vK_t(\vK_t + \lambda \vI_t)^{-1}\right)} \label{eq:C_enters} \\
	&\qquad \leq C \lambda^{-1/2}\sigma_{t}(\vx), \label{eq:bigstep9}
\end{align}
where:
\begin{itemize}[leftmargin=5ex,itemsep=0ex,topsep=0.25ex]
    \item Eq.~\eqref{eq:bigstep2} follows from the standard identity (see, e.g.,\cite[Eq.~(12)]{chowdhury17kernelized})
        \begin{equation}	
        	\label{eq:useful_matrix_identity}
        	\Phi_t^T(\Phi_t \Phi_t^T + \lambda \vI_t)^{-1} = (\Phi_t^T \Phi_t + \lambda \vI_{\Hc_k})^{-1}\Phi_t^T.
        \end{equation}
    \item Eq.~\eqref{eq:bigstep3} is by Cauchy-Schwartz.
    \item The first term $\lambda^{-1/2}\sigma_{t}(\vx)$ in \eqref{eq:variance_enters} follows from the following identity:
    \begin{equation}
    	\label{eq:variance_identity}
    	\sigma_{t}^2(\vx) = \lambda \phi(\vx)^T (\Phi_t^T \Phi_t + \lambda \vI_{\Hc_k})^{-1} \phi(\vx).
    \end{equation}
    To prove~\eqref{eq:variance_identity}, we first claim the following:
    \begin{equation}
    	\label{eq:phi_identity}
    	\phi(\vx) = \Phi_t^T \left(\Phi_t \Phi^T_t + \lambda \vI_t \right)^{-1} \Phi_t \phi(\vx) + \lambda \left(\Phi_t^T \Phi_t + \lambda \vI_{\Hc_k} \right)^{-1} \phi(\vx).
    \end{equation} 
    To see this, we apply~\eqref{eq:useful_matrix_identity} to the first term to obtain the equivalent expression
    \[
    	\phi(\vx) = \left(\Phi_t^T \Phi_t + \lambda \vI_{\Hc_k} \right)^{-1}\Phi_t^T \Phi_t \phi(\vx) + \lambda \left(\Phi_t^T \Phi_t + \lambda \vI_{\Hc_k} \right)^{-1} \phi(\vx).
    \]
    Multiplying from the left by $\left(\Phi_t^T \Phi_t + \lambda \vI_{\Hc_k} \right)$, we find that this is in turn equivalent to
    \[
    	\left(\Phi_t^T \Phi_t + \lambda \vI_{\Hc_k} \right) \phi(\vx) = \Phi_t^T \Phi_t \phi(\vx) + \lambda \phi(\vx),
    \]
    which trivially holds.  Then, note by the definition of $\sigma_t^2(\vx)$ and \eqref{eq:phi_identity} that
    \begin{align*}
    	\sigma_t^2(\vx) &= k(\vx,\vx) - k_t(\vx)^T \left(\vK_t +  \lambda \vI_t \right)^{-1} k_t(\vx) \\
    					&= \phi(\vx)^{T}\phi(\vx) - \phi(\vx)^T\Phi_t^T \left(\Phi_t \Phi^T_t + \lambda \vI_t \right)^{-1} \Phi_t \phi(\vx) \\
    					&\stackrel{\eqref{eq:phi_identity}}{=} \phi(\vx)^T\Phi_t^T \left(\Phi_t \Phi_t^T + \lambda \vI_{t} \right)^{-1} \Phi_t \phi(\vx) +  
    					\lambda \phi(\vx)^T (\Phi_t^T \Phi_t + \lambda\vI_{\Hc_k})^{-1} \phi(\vx) \\ 
    					&\quad -\phi(\vx)^T\Phi_t^T \left(\Phi_t \Phi_t^T + \lambda \vI_{t} \right)^{-1} \Phi_t \phi(\vx) \\
    					&= \lambda \phi(\vx)^T (\Phi_t^T \Phi_t + \lambda\vI_{\Hc_k})^{-1} \phi(\vx),
    \end{align*}
    yielding \eqref{eq:variance_identity}.
    The second term in \eqref{eq:variance_enters}  (i.e., the square root) follows by again applying \eqref{eq:useful_matrix_identity}.
    \item In \eqref{eq:max_ev_enters}, $\lambda_{\max}\left(\vK_t(\vK_t + \lambda \vI_t)^{-1}\right)$ denotes the largest eigenvalue of $\vK_t(\vK_t + \lambda \vI_t)^{-1}$. 
    \item Eq.~\eqref{eq:C_enters} follows since $\|\vc_t\|_{1} \leq C$ (see \eqref{eq:total_corruption}), and since the $\ell_1$ norm is always an upper bound on the $\ell_2$-norm.
    \item Eq.~\eqref{eq:bigstep9} follows since 
        $$\lambda_{\text{max}}(\vK_t(\vK_t + \lambda \vI_t)^{-1}) \leq 1.$$ 
    This follows since all eigenvectors of $\vK_t$ are also eigenvectors of $(\vK_t + \lambda \vI_t)^{-1}$, and hence, the eigenvalues of $\vK_t(\vK_t + \lambda \vI_t)^{-1}$ are of the form $\frac{\lambda (\vK_t)}{\lambda (\vK_t) + \lambda}$. Since, $\lambda (\vK_t)\geq 0$ and $\lambda > 0$, all the eigenvalues of $\vK_t(\vK_t + \lambda \vI_t)^{-1}$ are bounded by $1$.  
\end{itemize}

\section{Proof of Lemma~\ref{lem:conf_to_cumulative} (Regret Bound with Known Corruption)} \label{sec:pf_known_C}

Conditioned on the confidence bounds \eqref{eq:conf_bounds} being valid according to Lemma~\ref{conf_lemma_corrupted_known_C}, we have
\begin{align}
	&f(\vx^*) - f(\vx_t) \nonumber \\
        &\leq f(\vx^*) - \tilde{\mu}_{t-1} (\vx_t) + \beta_t \sigma_{t-1}(\vx_t) + \lambda^{-1/2}C \sigma_{t-1}(\vx_t) \label{eq:knownstep1} \\
		&\leq \tilde{\mu}_{t-1} (\vx^*) + \lambda^{-1/2}C \sigma_{t-1}(\vx^*) + \beta_t \sigma_{t-1}(\vx^*) - \tilde{\mu}_{t-1} (\vx_t) + \beta_t \sigma_{t-1}(\vx_t) + \lambda^{-1/2}C \sigma_{t-1}(\vx_t) \label{eq:knownstep1a} \\
		&\leq \tilde{\mu}_{t-1} (\vx_t) + \lambda^{-1/2}C \sigma_{t-1}(\vx_t) + \beta_t \sigma_{t-1}(\vx_t) - \tilde{\mu}_{t-1} (\vx_t) + \beta_t \sigma_{t-1}(\vx_t) + \lambda^{-1/2}C \sigma_{t-1}(\vx_t) \label{eq:knownstep2} \\
		&= 2(\lambda^{-1/2}C + \beta_t)\sigma_{t-1}(\vx_t). \label{eq:knownstep3}
\end{align}
where \eqref{eq:knownstep1} uses the lower confidence bound from \eqref{eq:conf_bounds}, \eqref{eq:knownstep1a} uses the upper confidence bound from \eqref{eq:conf_bounds},  and \eqref{eq:knownstep2} uses the selection rule in \eqref{eq:enlarged_ucb}.

When $\lambda \geq 1$, we have from~\cite[Lemma 4]{chowdhury17kernelized} that\footnote{The statement of \cite[Lemma 4]{chowdhury17kernelized} uses $\lambda = 1 + 2/T$, but the proof states the result for general $\lambda \ge 1$.}
\begin{equation}
	\sum_{t=1}^T \sigma_{t-1}(\vx_t) \leq \sqrt{4 T \lambda \gamma_T}. \label{eq:sigma_bound}
\end{equation}
This is a variant of a more widely-used upper bound on $\sum_{t=1}^T \sigma_{t-1}(\vx_t)$ in terms of $\gamma_T$ from \cite{srinivas2009gaussian}. 

We set $\lambda = 1$ in accordance with the lemma statement, and sum over the time steps:
\begin{align}
	R_T
    & = \sum_{t=1}^T \big( f(\vx^*) - f(\vx_t) \big) \\
    &\leq (2C + 2\beta_T) \sum_{t=1}^T \sigma_{t-1}(\vx_t) \label{eq:RTstep1} \\
	&\leq \left(2C + 2\beta_T\right) \sqrt{4T \gamma_T} \label{eq:RTstep2} \\
	&\leq \left(2C + 2B + 2\sigma \sqrt{2\left(\gamma_{T} + \ln (\tfrac{1}{\delta})\right)} \right) \sqrt{4T \gamma_T}, \label{eq:RTstep4}
\end{align}
where \eqref{eq:RTstep1} uses \eqref{eq:knownstep3} and the monotonicity of $\beta_t$, \eqref{eq:RTstep2} uses \eqref{eq:sigma_bound}, and \eqref{eq:RTstep4} substitutes the choice of $\beta_t$ in \eqref{eq:standard_beta} and applies $\gamma_{T-1} \le \gamma_T$.  Hence, we have $R_T = \mathcal{O}\big( (B + C + \sqrt{\ln(1 / \delta)})\sqrt{\gamma_T T} + \gamma_{T}\sqrt{T}\big)$, which establishes the lemma.

\section{Bounding the Simple Regret} \label{sec:simple}

While we have focused exclusively on the cumulative regret in our exposition, we can easily adapt our analysis to handle the simple regret similarly to the idea used in the proof of \cite[Theorem 1]{bogunovic2018adversarially}. We outline this procedure for Theorem \ref{thm:known_C_thm}, since all of the other results can be adapted in the same manner.

We claim that under the setup of Theorem \ref{thm:known_C_thm}, for a given $\Delta > 0$, Algorithm \ref{alg:ucb_with_enlarged_conf_bounds} achieves $f(\vx^*) - f(\vx^{(T)}) \leq \Delta$ after $T = \mathcal{O}\left( \frac{\gamma_T (\beta_T + C)^2}{\Delta^2} \right)$ rounds, where the reported point $\vx^{(T)}$ is defined as
\begin{equation}\label{eq:simple_regret_reported_point}
	\vx^{(T)} = \vx_{t^*}, \quad \text{with} \quad t^* = \argmax_{1,\dots,T} \left \lbrace \tilde{\mu}_{t-1}(\vx_t) - (C + \beta_t)\sigma_{t-1}(\vx_t)\right \rbrace.
\end{equation}

To prove this claim, we continue from the end of Appendix \ref{sec:pf_known_C}. We set $\lambda=1$ as before, and define 
    $$\bar{r}(\vx_t):= f(\vx^*) - \tilde{\mu}_{t-1}(\vx_t) + (C + \beta_t)\sigma_{t-1}(\vx_t).$$
Using \eqref{eq:knownstep1}, we have $f(\vx^*) - f(\vx) = r(\vx_t)\leq \bar{r}(\vx_t)$ for each $t \geq 1$. From the definition of the reported point $\vx^{(T)}$ in~\eqref{eq:simple_regret_reported_point}, we have that $t^*$ is the time index with the smallest value of $\bar{r}(\vx_t)$.  It follows that
\begin{align}
\bar{r}(\vx^{(T)}) 
    &\leq \frac{1}{T} \sum_{t=1}^T 2(C + \beta_t) \sigma_{t-1}(\vx_t) \label{eq:barstep1} \\
    &\leq \frac{2(C + \beta_T)}{T} \sum_{t=1}^T \sigma_{t-1}(\vx_t) \label{eq:barstep2} \\
    &\leq \frac{2(C + \beta_T)}{T} \sqrt{4T \gamma_T}, \label{eq:barstep3}
\end{align}
where \eqref{eq:barstep1} upper bounds the minimum by the average, \eqref{eq:barstep2} uses the monotonicity of $\beta_t$, and \eqref{eq:barstep3} uses \eqref{eq:sigma_bound} with $\lambda = 1$. 

Re-arranging \eqref{eq:barstep3}, we find that after $T = \mathcal{O}\big( \frac{\gamma_T (\beta_T + C)^2}{\Delta^2} \big)$ time steps, $\bar{r}(\vx^{(T)}) \le \Delta$, which further implies that $r(\vx^{(T)}) \leq \Delta$.

\section{Proof of Lemma~\ref{lemma:corruption_observed_at_S} (Total Corruption Observed by $S$)}

We follow the proof of \cite[Lemma 3.3]{lykouris2018stochastic}, making use of the following martingale concentration inequality.\footnote{This result is presented in \cite{beygelzimer2011contextual} for the filtration $\Fc_t$ generated by $M_1,\dotsc,M_T$ itself, but the proof applies in the general case.  To prove Lemma \ref{lemma:corruption_observed_at_S}, we could in fact resort to the classical martingale concentration bound of Freedman \cite{freedman1975tail}, but we found the form given in \cite{beygelzimer2011contextual} to be more convenient.}

\begin{lemma}{\em \cite[Lemma 1]{beygelzimer2011contextual}}
\label{lemma:martingale_conc_ineq}
Let $M_1, \dots, M_T$ be a sequence of real-valued random variables forming a martingale with respect to a filtration $\{\Fc_t\}$, i.e., $\E[M_t|\Fc_{t-1}] = 0$, and suppose that $M_t \leq R$ almost surely. Then for any $\delta > 0$, the following holds:
\[
	\P \bigg[\sum_{t=1}^T M_t  \leq \frac{V}{R}(e-2) + R \ln (1/ \delta) \bigg] \geq 1 - \delta,
\] 
where $V = \sum_{t=1}^T  \E [M^2_t|\Fc_{t-1}]$.
\end{lemma}

Let $\vx_t^{(S)}$ be the point that would be selected at time $t$ if instance $S$ were chosen. We let $C_t = |c_t(\vx_t^{(S)})| \bone\{ A_t = S \}$ denote the amount of corruption observed by instance $S$ at time $t$ in Algorithm~\ref{alg:universal_alg_given_C}. 

Let $\Hc_{t-1}$ denote the history (i.e., all selected instances $A_i \in \{F,S\}$, inputs $\vx_i \in D$, and observations $\tilde{y}_i \in \mathbb{R}$) prior to round $t$.  Noting that $\vx_t^{(S)}$ is deterministic given $\Hc_{t-1}$, we find that $C_t$ is a random variable equaling $|c_t(\vx_t^{(S)})|$ with probability $\rho:=\min\lbrace 1,C^{-1}\rbrace$ and $0$ otherwise.  As a result, we can define the following martingale sequence:
\[
	M_t = C_t - \E[C_t| \Hc_{t-1} ],
\]
where $\E[C_t| \Hc_{t-1} ] = \rho |c_t(\vx_t^{(S)})|$ as stated above. Since $c_t(\vx)\in [-B_0,B_0]$ for all $t$ and $\vx \in D$ (see Section \ref{sec:problem}), we have $M_t \leq B_0$ for all $t$. Hence, we can set $R = B_0$ in Lemma~\ref{lemma:martingale_conc_ineq}.

Next, we note the following: 	
\begin{align*}
	\E [M^2_t|\Hc_{t-1}] &= \rho \left( |c_t(\vx_t^{(S)})| - \rho |c_t(\vx_t^{(S)})| \right)^2 + (1-\rho)\left(\rho |c_t(\vx_t^{(S)})| \right)^2 \\
	&= \rho c_t(\vx_t^{(S)})^2(1-\rho)^2 + (1-\rho)(\rho c_t(\vx_t^{(S)}))^2\\
	&\leq \rho c_t(\vx_t^{(S)})^2 + \rho c_t(\vx_t^{(S)})^2 \\
	&= 2\rho c_t(\vx_t^{(S)})^2 \\
    & \leq 2\rho B_0|c_t(\vx_t^{(S)})|.
\end{align*}
where the two inequalities use $\rho \in [0,1]$ and $c_t(\vx_t^{(S)}) \leq B_0$ respectively. By summing over all the rounds and using the definition of $C$ in \eqref{eq:total_corruption}, we obtain
\begin{align*}
	V = \sum_{t=1}^T \E [M^2_t|\Hc_{t-1}] &\leq 2 B_0 \rho \sum_{t=1}^{T} |c_t(\vx_t^{(S)})|\leq 2 B_0 \rho C \leq 2B_0,
\end{align*}
since $\rho \le C^{-1}$.  Applying Lemma~\ref{lemma:martingale_conc_ineq}, we have with probability at least $1-\delta$ that
\begin{equation}
\sum_{t=1}^{T} M_t \leq \frac{2B_0}{B_0}(e-2) + B_0\ln (1/\delta) \leq 2 + B_0\ln (1/\delta). \label{eq:sum_M_bound}
\end{equation}
Finally, we complete the proof of Lemma~\ref{lemma:corruption_observed_at_S} by adding the total expected corruption:
\begin{align*}
\sum_{t=1}^T C_t &= \sum_{t=1}^{T} M_t + \sum_{t=1}^T \E \left[C_t| \Hc_{t-1}\right] \\
&\leq  3 + B_0\ln (1/\delta),
\end{align*}
where we have used \eqref{eq:sum_M_bound} and $\sum_{t=1}^T \E \left[C_t| \Hc_{t-1}\right] = \rho \sum_{t=1}^T |c_t(\vx_t^{(S)})| \le \rho C \le 1$.

\section{Proof of Lemma~\ref{lemma:F_queries_no_S_suboptimal_point} (Characterizing the Points Not Sampled by $F$)}

Consider any round $t \in \lbrace 1,\dots, T \rbrace$ and any point $\vx \in \Sc_t$ (see \eqref{eq:def_St}). We wish to show that $F$ never selects $\vx$, i.e., $\vx_t \neq \vx$. To establish this, it suffices to prove that
	\begin{equation}
		\label{eq:lemma_to_prove}
		\min_{A \in \lbrace F,S \rbrace} \ucbabar(\vx; 1) < \min_{A \in \lbrace F,S \rbrace} \ucbabar(\vx'; 1). 
	\end{equation}
	for some $\vx' \in D$; this means that $\vx'$ is favored over $\vx$ according to the selection rule of $F$.

	To show \eqref{eq:lemma_to_prove}, we first trivially write
		\begin{equation}
			\label{eq:lemma_aux_1}
			\min_{A \in \lbrace F,S \rbrace} \ucbabar (\vx; 1) \leq \ucbsbar (\vx; 1).
		\end{equation}
	Since $\vx \in \Sc_t$, by the definition of $\Sc_t$ in \eqref{eq:def_St}, there exists $\vx' \in D$ such that
		\begin{equation}
		\label{eq:lemma_aux_2}
			\ucbsbar (\vx; 1) < \lcbsbar(\vx';1).
		\end{equation}
	Moreover, the following two equations provide upper bounds on $\lcbsbar(\vx'; 1)$:
		\begin{align}
			\lcbsbar(\vx'; 1) &\leq \ucbsbar(\vx';1) \label{eq:lcbs_bound_1}\\
			\lcbsbar(\vx'; 1) &\leq \ucbfbar(\vx';1), \label{eq:lcbs_bound_2}
		\end{align}
	where \eqref{eq:lcbs_bound_1} follows from the validity of the confidence bounds (see \eqref{eq:confidence_alg2_2}), and \eqref{eq:lcbs_bound_2} is due to $A_t = F$, which means that the condition~\eqref{eq:condition} used in \fsucb~(Line 12) is not satisfied and thus it cannot hold that $\lcbsbar(\vx'; 1) > \ucbfbar(\vx';1)$.  

	From \eqref{eq:lcbs_bound_1} and \eqref{eq:lcbs_bound_2} we have $\lcbsbar(\vx'; 1) \leq \min_{\lbrace F,S \rbrace} \ucbabar(\vx';1)$, and from~\eqref{eq:lemma_aux_1} and \eqref{eq:lemma_aux_2} we have 
	$\lcbsbar(\vx'; 1) > \min_{A \in \lbrace F,S \rbrace} \ucbabar(\vx;1),$
	which together prove that \eqref{eq:lemma_to_prove} holds.

\section{Proof of Lemma~\ref{lemma:time_to_suboptimality} (Characterizing the Points Ruled Out via $S$)}

Although we consider the $S$ instance run with $\alpha = 2$, we are interested in how long it takes before the following (corresponding to $\alpha = 1$) is observed for the given suboptimal~$\vx$ and some $\vx' \in D$:
\begin{equation} \label{eq:S_t_condition}
    \ucbs(\vx;1) < \lcbs(\vx';1).
\end{equation}
Since $\ucbsbar$ and $\lcbsbar$ are tighter confidence bounds than $\ucbs$ and $\lcbs$, \eqref{eq:S_t_condition} holding implies that 
\begin{equation} \label{eq:S_t_condition2}
    \ucbsbar(\vx;1) < \lcbsbar(\vx';1),
\end{equation}
meaning that $\vx \in \Sc_{t_S}$ (see  \eqref{eq:def_St}). Since $\ucbsbar$ and $\lcbsbar$ are monotone, \eqref{eq:S_t_condition2} holding for some $t_S$ means that it continues to hold for all $t'_S > t_S$.  Hence, to establish the lemma, it suffices to show that after $t_S$ rounds (with $t_S$ given in \eqref{eq:time_S}), there exists a point $\vx' \in D$ such that \eqref{eq:S_t_condition} holds.

Since this proof only concerns points selected by $S$, we abuse notation slightly and let $\vx_i$ denote the $i$-th point queried by $S$.  We use the fact that the instant regret incurred by the $S$ instance satisfies
\begin{equation}
    r(\vx_{i})=f(\vx^*) - f(\vx_{i}) \leq 2 \alpha \beta_{i}^{(S)} \sigma_{i-1}^{(S)}(\vx_{i}) \label{eq:standard1}
\end{equation}
(via an identical argument\footnote{See also \eqref{eq:instant_regret_S} in Appendix \ref{sec:corr_stoch_pf}.} to \eqref{eq:knownstep3}), and the sum of posterior standard deviations satisfies 
\begin{equation}
    \frac{1}{t_S} \sum_{i=1}^{t_S} \sigma_{i-1}^{(S)}(\vx_i) \le \sqrt{\frac{4\gamma_{t_{S}}}{t_S}} \label{eq:standard2}
\end{equation}
when we set $\lambda=1$ (by a direct application of \eqref{eq:sigma_bound}). Combining these gives
    \begin{equation}
        \frac{1}{t_S} \sum_{i=1}^{t_S} r(\vx_i) \le \frac{1}{t_S} \sum_{i=1}^{t_S} 2\alpha \beta_i^{(S)} \sigma^{(S)}_{i-1}(\vx_i)  \le \sqrt{ \frac{C_1 (\beta^{(S)}_{t_S})^{2} \gamma_{t_S}}{t_S}}, \label{eq:r_sum}
    \end{equation}
    where $C_1 = 16\alpha^2$.
It is useful to ``invert'' the right-hand side of \eqref{eq:r_sum}; to do this, we define the function
\begin{equation}
    \tau(\Delta) = \min\bigg\{ \tau \,:\, \sqrt{ \frac{C_1 (\beta^{(S)}_{\tau})^{2} \gamma_{\tau}}{\tau}} \le \Delta \bigg\}. \label{eq:tau}
\end{equation}

Since \eqref{eq:r_sum} and \eqref{eq:tau} state that the ``average'' value of $2\alpha \beta_i^{(S)} \sigma^{(S)}_{i-1}(\vx_i)$ by time $\tau(\Delta)$ is at most $\Delta$, we deduce that
\begin{equation}
    \forall \Delta > 0, \exists i \le \tau(\Delta) \text{ such that } 2\alpha \beta_i^{(S)} \sigma^{(S)}_{i-1}(\vx_i) \le \Delta. \label{eq:Delta_init}
\end{equation}
That is, at least one time index $i$ yields a value less than or equal to the average.

Now consider the given $\vx \in D$ with instant regret satisfying $r(\vx) \ge \Delta_0 > 0$ in accordance with the lemma statement.  Setting the parameter $\Delta = \frac{\Delta_0}{10}$ in \eqref{eq:Delta_init} gives 
\begin{align}
    \exists i \le \tau(\Delta_0/10)  
        &\text{ such that }  2\alpha \beta_i^{(S)} \sigma^{(S)}_{i-1}(\vx_i) \le \frac{\Delta_0}{10} \label{eq:i_bound1} \\
        &\text{ and hence }  r(\vx_i) \le \frac{\Delta_0}{10}, \label{eq:i_bound2}
\end{align}
where \eqref{eq:i_bound2} follows from \eqref{eq:standard1}.  This means that $\vx_i$ is much closer to optimal than $\vx$ is.  The properties in \eqref{eq:i_bound1} and \eqref{eq:i_bound2} allow us to characterize the confidence bounds of $\vx_i$:
\begin{align}
    \ucb^{(S)}_{i-1}(\vx_i;\alpha) 
        &= \lcb^{(S)}_{i-1}(\vx_i;\alpha) + 2\alpha\beta^{(S)}_i \sigma^{(S)}_{i-1}(\vx_i) \label{eq:i_prop1_step1} \\
        &\le f(\vx_i) + 2\alpha\beta^{(S)}_i \sigma^{(S)}_{i-1}(\vx_i) \label{eq:i_prop1_step2} \\
        &\le  f(\vx^*) + \frac{\Delta_0}{10}, \label{eq:i_prop1}
\end{align}
where \eqref{eq:i_prop1_step1} uses the definition of the confidence bounds in \eqref{eq:ucbA}--\eqref{eq:lcbA}, \eqref{eq:i_prop1_step2} uses the validity of the confidence bounds in \eqref{eq:confidence_alg2_2}, and \eqref{eq:i_prop1} uses \eqref{eq:i_bound1}.  Similarly,
\begin{align}
    \lcb^{(S)}_{i-1}(\vx_i;\alpha) 
    &= \ucb^{(S)}_{i-1}(\vx_i;\alpha) - 2\alpha\beta^{(S)}_i \sigma^{(S)}_{i-1}(\vx_i)  \label{eq:i_prop2_step0} \\
    &\ge \ucb^{(S)}_{i-1}(\vx^*;\alpha) - 2\alpha\beta^{(S)}_i \sigma^{(S)}_{i-1}(\vx_i) \label{eq:i_prop2_step1} \\
    &\ge  f(\vx^*) - 2\alpha\beta^{(S)}_i \sigma^{(S)}_{i-1}(\vx_i)  \label{eq:i_prop2_step2} \\
    &\ge f(\vx^*) - \frac{\Delta_0}{10}, \label{eq:i_prop2}
\end{align}	
where \eqref{eq:i_prop2_step0} is the same as \eqref{eq:i_prop1_step1}, \eqref{eq:i_prop2_step1} uses the UCB selection rule, \eqref{eq:i_prop2_step2} uses the validity of the confidence bounds, and \eqref{eq:i_prop2} uses \eqref{eq:i_bound1}.  Combining \eqref{eq:i_prop1} and \eqref{eq:i_prop2}, we find that the confidence interval $[\lcb^{(S)}_{i-1}(\vx_i;\alpha), \ucb^{(S)}_{i-1}(\vx_i;\alpha)]$ is within the range
\begin{equation}
    \mathcal{I} = \Big[f(\vx^*) - \frac{\Delta_0}{10}, f(\vx^*) + \frac{\Delta_0}{10}\Big]. \label{eq:calI}
\end{equation}

It also holds that  
\begin{equation}
    \ucb^{(S)}_{i-1}(\vx;\alpha) \le \ucb^{(S)}_{i-1}(\vx_i;\alpha) \label{eq:also_ucb}
\end{equation}
by the UCB rule used in the $S$ instance.  For this fixed $i \leq \tau(\Delta_0 / 10)$ and $\vx_i$, there are then two possible cases that we need to consider:
\begin{enumerate}[leftmargin=5ex,itemsep=0ex,topsep=0.25ex]
    \item If it also holds that $\ucb^{(S)}_{i-1}(\vx;\alpha) < \lcb^{(S)}_{i-1}(\vx_i;\alpha)$, then we immediately obtain 
        $$\ucb^{(S)}_{i-1}(\vx;1) < \lcb^{(S)}_{i-1}(\vx_i;1)$$ 
    because we chose $\alpha = 2$, and decreasing $\alpha$ only makes $\ucb^{(S)}_{i-1}(\cdot;\alpha)$  decrease and $\lcb^{(S)}_{i-1}(\cdot;\alpha)$ increase (see \eqref{eq:ucbA}--\eqref{eq:lcbA}). Hence, the condition in~\eqref{eq:S_t_condition} holds as required. 
    \item Otherwise, by \eqref{eq:also_ucb}, we must have $$\lcb^{(S)}_{i-1}(\vx_i,\alpha) \le \ucb^{(S)}_{i-1}(\vx;\alpha) \le \ucb^{(S)}_{i-1}(\vx_i;\alpha).$$  By~\eqref{eq:i_prop1} and~\eqref{eq:i_prop2}, this means that $\ucb^{(S)}_{i-1}(\vx;\alpha)$ lies in the interval $\Ic$ given in \eqref{eq:calI}.

    Since the confidence bounds \eqref{eq:confidence_alg2_2} are valid and $f(\vx) \le f(\vx^*) - \Delta_0$ (i.e., $r(\vx) \ge \Delta_0$), we must also have $\lcb^{(S)}_{i-1}(\vx;\alpha) \le f(\vx^*) - \Delta_0$.  Comparing this with $\Ic$ above, we notice a gap of at least $\frac{9\Delta_0}{10}$ between the upper and lower confidence bounds at $\vx$.  Let this gap be denoted by ${\rm Gap}(\alpha) \ge \frac{9\Delta_0}{10}$.
    
    The confidence bounds $\ucb_{i-1}^{(S)}(\vx;\alpha)$ and $\lcb_{i-1}^{(S)}(\vx;\alpha)$ are equal to $\tmu \pm \frac{1}{2} {\rm Gap}(\alpha)$, where $\tmu$ is shorthand for the corrupted posterior mean.  When we compare to $\ucb^{(S)}_{i-1}(\vx;1)$ and $\lcb^{(S)}_{i-1}(\vx;1)$, the value $\tmu$ remains unchanged, but we have ${\rm Gap}(1) = \frac{1}{\alpha} {\rm Gap}(\alpha)$; see \eqref{eq:ucbA}--\eqref{eq:lcbA}.  Therefore, we have
    \begin{align*} 
        \ucb^{(S)}_{i-1}(\vx;1)
            &= \ucb^{(S)}_{i-1}(\vx;\alpha) - \frac{1}{2} \Big(1 - \frac{1}{\alpha}\Big) {\rm Gap}(\alpha) \\
            &\le \ucb^{(S)}_{i-1}(\vx;\alpha) - \frac{1}{2} \Big(1 - \frac{1}{\alpha}\Big)  \frac{9\Delta_0}{10}
    \end{align*}
    since ${\rm Gap}(\alpha) \ge \frac{9\Delta_0}{10}$.
    Substituting $\alpha = 2$ gives $\ucb^{(S)}_{i-1}(\vx;1) \le \ucb^{(S)}_{i-1}(\vx;2) - \frac{9\Delta_0}{40}$.  Since the width of the interval $\Ic$ (in which $\ucb^{(S)}_{i-1}(\vx;2)$ lies) is only $\frac{2\Delta_0}{10} = \frac{8\Delta_0}{40}$, we conclude that $\ucb^{(S)}_{i-1}(\vx;1)$ lies strictly below $\Ic$.
    
    On the other hand, using \eqref{eq:i_prop1} and \eqref{eq:i_prop2}, we see that the entire confidence interval for $\vx_i$ lies within $\Ic$ (recall that replacing $\alpha > 1$ by $\alpha = 1$ only shrinks this interval).  Hence, $\ucb^{(S)}_{i-1}(\vx;1) < \lcb^{(S)}_{i-1}(\vx_i;1)$, as required.
\end{enumerate}
Recall that the above findings all correspond to some time index $i \le \tau(\Delta_0/10)$. Hence, \eqref{eq:time_S} follows by setting $t_S = \tau(\Delta_0/10)$.

\section{Proof of Theorem~\ref{thm:corrupted_vs_stochastic_thm} (Regret Bound in the Known-or-Zero Setting)} \label{sec:corr_stoch_pf}

Throughout the proof, we condition on the events~\eqref{eq:confidence_alg2_1}--\eqref{eq:confidence_alg2_3} that simultaneously hold with probability at least $1 - \frac{4\delta}{5}$.

\subsection{Non-corrupted case}

Recall that at time $t$, the chosen instance and input are denoted by $A_t$ and $\vx_t$, respectively, and we use $t_A$ to denote the number of times an instance $A \in \{F,S\}$ has been chosen up to time~$t$. 

In the non-corrupted case, the condition~\eqref{eq:condition} cannot hold (conditioned on the events~\eqref{eq:confidence_alg2_1} and \eqref{eq:confidence_alg2_2}), since the confidence bounds for both $S$ and $F$ are valid and hence $\ucbft(\vx;1)$ can never be smaller than $\lcbst(\vx;1)$. Consequently, Algorithm~\ref{alg:universal_alg_given_C} selects only $S$ or $F$, and never switches permanently to Algorithm~\ref{alg:ucb_with_enlarged_conf_bounds}. 

First, we consider the case that $A_t = S$ is used to select $\vx_t$ for some $t$. We have
\begin{align}
	f(\vx^*) - f(\vx_t) &\leq \ucbs(\vx^*; \alpha) - f(\vx_t) \label{eq:caseS_1} \\
						&\leq \ucbs(\vx_t; \alpha) - f(\vx_t) \label{eq:S_rule_used} \\ 
						&\leq \ucbs(\vx_t; \alpha) - \lcbs(\vx_t;\alpha)  \label{eq:caseS_3} \\
						&\leq 2 \alpha \beta_{t_S}^{(S)} \sigma_{t_{S}-1}^{(S)}(\vx_t), \label{eq:instant_regret_S}
\end{align}
where \eqref{eq:caseS_1} and \eqref{eq:caseS_3} use the validity of the confidence bounds, \eqref{eq:S_rule_used} follows from the selection rule of $S$, and \eqref{eq:instant_regret_S} uses the definitions \eqref{eq:ucbA}--\eqref{eq:lcbA}.

Next, we consider the case that $A_t = F$ is used to select $\vx_t$ for some $t$. We have
\begin{align}
	f(\vx^*) - f(\vx_t) &\leq \min_{A \in \lbrace F,S \rbrace} \ucbabar(\vx^*; 1) - f(\vx_t) \label{eq:caseF_1} \\
	&\leq \min_{A \in \lbrace F,S \rbrace} \ucbabar(\vx_t; 1) - f(\vx_t) \label{eq:caseF_2} \\
	&\leq \ucbfbar(\vx_t; 1) - f(\vx_t) \label{eq:caseF_3} \\
	&\leq \ucbfbar(\vx_t; 1) - \lcbfbar(\vx_t;1) \label{eq:caseF_4} \\
	&\leq 2\beta_{t_{F}}^{(F)} \sigma_{t_{F}-1}^{(F)}(\vx_t), \label{eq:instant_regret_F}
\end{align}
where~\eqref{eq:caseF_1} and \eqref{eq:caseF_4} use the validity of the confidence bounds, \eqref{eq:caseF_2} uses the selection rule of $F$, and \eqref{eq:instant_regret_F} follows similarly to \eqref{eq:instant_regret_S} by noting that the intersected confidence bounds are at least as tight as the non-intersected ones.

The regret $R_T$ of Algorithm~\ref{alg:universal_alg_given_C} after $T$ rounds can be trivially bounded by the sum $R^{(S)}_T + R^{(F)}_T$, where $R^{(A)}_T$ is the regret of instance $A$ when run for $T$ rounds in the non-corrupted case:
\begin{align}
	R_T &\leq R_T^{(F)} + R_T^{(S)} \\
		&\leq \sum_{t_F=1}^{T} 2\beta_{t_{F}}^{(F)} \sigma_{t_{F}-1}^{(F)}(\vx_{t_F}) + \sum_{t_S=1}^{T} 2 \alpha \beta_{t_S}^{(S)} \sigma_{t_{S}-1}^{(S)}(\vx_{t_S}) \label{eq:regret_sum} \\
		&\leq 2\beta_{T}^{(F)} \sum_{t_F=1}^{T} \sigma_{t_{F}-1}^{(F)}(\vx_{t_F}) + 2 \alpha \beta_{T}^{(S)} \sum_{t_S=1}^{T}\sigma_{t_{S}-1}^{(S)}(\vx_{t_S}) \label{eq:monotonicity_of_beta_S_and_F} \\
		&\leq 2\beta_{T}^{(F)} \sqrt{4T\gamma_T} + 2 \alpha \beta_{T}^{(S)} \sqrt{4T\gamma_T} \label{eq:mmi_bound_enters} \\
		&\leq 4 \alpha \beta_{T}^{(S)} \sqrt{4T\gamma_T} \label{eq:beta_S_larger_than_beta_F},
\end{align}
where~\eqref{eq:regret_sum} follows from~\eqref{eq:instant_regret_S} and~\eqref{eq:instant_regret_F}, \eqref{eq:monotonicity_of_beta_S_and_F} follows since both $\beta^{(S)}_{t_S}$ and $\beta^{(F)}_{t_F}$ are non-decreasing in the time index, \eqref{eq:mmi_bound_enters} follows from \eqref{eq:sigma_bound} by setting $\lambda=1$,
and~\eqref{eq:beta_S_larger_than_beta_F} follows since $\alpha \ge 1$ and  $\beta^{(S)}_T \geq \beta^{(F)}_T$ (see \eqref{eq:beta_F}--\eqref{eq:beta_S}).

Substituting $\beta^{(S)}_{T} = B + \sigma \sqrt{2\left(\gamma_{T - 1} + \ln \left( \frac{5}{\delta} \right)\right)} +  (3 + B_0 \ln \left( \frac{5}{\delta} \right))$  and $\alpha = 2$ in~\eqref{eq:beta_S_larger_than_beta_F}, we arrive at the regret bound, i.e., with probability at least $1 - \frac{4}{5}\delta \ge 1-\delta$, the regret of Algorithm~\ref{alg:universal_alg_given_C} after $T$ rounds is
    $$R_T = \mathcal{O}\left(\Big(B + B_0\ln (\tfrac{1}{\delta}) + \sqrt{\ln (\tfrac{1}{\delta})}\Big) \sqrt{T\gamma_T} + \gamma_T \sqrt{T}\right).$$

\subsection{$C$-corrupted case}
\label{sec:corrupted_case}

Similarly to the non-corrupted case, we condition on \eqref{eq:confidence_alg2_1}--\eqref{eq:confidence_alg2_3}, and we set $\lambda=1$.
We first address the two parts of \fsucb~whose contributions to the cumulative regret are the simplest to handle: That from Algorithm~\ref{alg:ucb_with_enlarged_conf_bounds}, and that from the slow instance $S$.

Supposing that Algorithm~\ref{alg:ucb_with_enlarged_conf_bounds} is run for $T' \le T$ rounds, we simply use the confidence bounds \eqref{eq:confidence_alg2_3} and apply Lemma~\ref{lem:conf_to_cumulative} (with $\frac{\delta}{5}$ in place of $\delta$): If
 	$\beta_{t}^{(A_1)} = B + \sigma \sqrt{2\left(\gamma_{t - 1} + \ln \left( \tfrac{5}{\delta} \right)\right)} + C$, then the cumulative regret after $T'$ rounds satisfies
\begin{equation} \label{eq:corruted_A1_regret}
	R_{T'}^{(A_1)} = \mathcal{O}\left(\Big(B + C + \sqrt{\ln(\tfrac{1}{\delta})}\Big) \sqrt{\gamma_{T'} T'} + \gamma_{T'}\sqrt{T'}\right). 
\end{equation} 

The regret obtained by $S$ is analyzed in the same way via Lemma~\ref{lem:conf_to_cumulative}, but with $B_0 \ln \big( \tfrac{5}{\delta} \big)$ in place of $C$, and the confidence bounds \eqref{eq:confidence_alg2_2} in place of \eqref{eq:confidence_alg2_3}.  Lemma \ref{lem:conf_to_cumulative} then implies that the regret coming from $S$ for a total of $T'$ rounds satisfies
\begin{equation} \label{eq:corrupted_S_regret}
	R_{T'}^{(S)} = \mathcal{O}\left(\Big(B + B_0\ln (\tfrac{1}{\delta}) + \sqrt{\ln (\tfrac{1}{\delta})}\Big) \sqrt{\gamma_{T'}T'} + \gamma_{T'} \sqrt{T'}\right), 
\end{equation}
which is the same as \eqref{eq:corruted_A1_regret} but with $B_0 \ln \big( \tfrac{5}{\delta} \big)$ in place of $C$ (and possibly a different $T'$ value).
It now only remains to bound the regret of the $F$ instance in the corrupted case.

\textbf{Regret incurred by the $F$ instance.}
First, we recall a few facts. The $F$-confidence bounds in \eqref{eq:confidence_alg2_1} are only valid when there is no corruption, and hence they cannot be used to characterize the regret of the $F$ instance in the corrupted case. Unlike the $F$-confidence bounds, the $S$-confidence bounds in \eqref{eq:confidence_alg2_2} are valid even in the corrupted case, and they are useful since the $F$ rule explicitly depends on them (\fsucb, Line 6). In Lemma~\ref{lemma:F_queries_no_S_suboptimal_point}, we have shown that no point that is suboptimal according to the $S$-confidence bounds is sampled by the $F$ instance. Subsequently, in Lemma~\ref{lemma:time_to_suboptimality}, we have characterized how many points need to be queried in $S$ before this occurs.  More formally, the results of Lemmas~\ref{lemma:F_queries_no_S_suboptimal_point} and~\ref{lemma:time_to_suboptimality} (with $\alpha=2$) state that by time 
\begin{equation} \label{eq:t_S_app}
	t_S = \min\big\{ \tau \,:\, 8\beta^{(S)}_{\tau}\sqrt{ \tfrac{\gamma_{\tau}}{\tau}} \le \tfrac{\Delta_0}{10} \big\},
\end{equation}
 all $\Delta_0$-suboptimal points are ruled out and are not sampled by $F$ in the subsequent time steps.  We observe that the following two statements are equivalent:
\begin{itemize}[leftmargin=5ex,itemsep=0ex,topsep=0.25ex]
    \item After time $t_S = \min\big\{ \tau \,:\, 8 \beta^{(S)}_{\tau}\sqrt{ \tfrac{ \gamma_{\tau}}{\tau}} \le \tfrac{\Delta_0}{10} \big\}$, the instant regret of each point selected by $F$ is at most $\Delta_0$;
    \item After time $t_S$, the instant regret of each point selected by $F$ is at most $80 \beta^{(S)}_{t_S}\sqrt{ \tfrac{\gamma_{t_S}}{t_S}}$.
\end{itemize}
This is by a simple inversion; if we set $\Delta_0 = 80 \beta^{(S)}_{\tau}\sqrt{ \tfrac{\gamma_{\tau}}{\tau}}$ in \eqref{eq:t_S_app} then it trivially holds that $8 \beta^{(S)}_{\tau}\sqrt{ \tfrac{ \gamma_{\tau}}{\tau}} \le \tfrac{\Delta_0}{10}$.

We now seek to characterize how many times $F$ is selected in between successive selections of $S$.  If $C \le 1$, then this is trivial, since $S$ is always selected, so in the following we focus on $C > 1$.  We will establish that with probability at least $1 - \frac{\delta}{5}$, in between any two selections of $S$ (or prior to the first such selection), there are at most $C \ln \frac{5T}{\delta}$ selections of $F$ with probability at least $1 - \frac{\delta}{5}$.  We henceforth denote this event by $\Ac$.

To establish the preceding claim, fix an integer $N > 0$, and observe that after any given selection of $S$, the probability of selecting $F$ for the next $N$ rounds is $\big(1 - \frac{1}{C}\big)^N \le e^{-N/C}$. Hence, if $N = C \ln \frac{1}{\delta'}$, then the probability is at most $\delta'$.  The number of selections of $S$ is trivially at most $T$, so taking a union bound over at most $T$ associated events, we obtain $\P[\Ac] \ge 1-\frac{\delta}{5}$ when $\delta' = \frac{\delta}{5T}$.

By the union bound, the event $\Ac$ and the events in~\eqref{eq:confidence_alg2_1}--\eqref{eq:confidence_alg2_3} hold simultaneously with probability at least $1 - \tfrac{4}{5} \delta - \tfrac{1}{5}\delta =  1-\delta$. Conditioned on these events, when \fsucb~is run for $T$ rounds, the cumulative regret of the points selected by $F$ satisfies\footnote{We could slightly improve this bound by replacing $\gamma_T$ by $\gamma_{\tfrac{T}{N}}$, but we proceed with the former since it is simpler and only slightly weaker.}
\begin{align}
    R_T^{(F)} 
        & \le 2B_0 N + N \cdot 80 \beta^{(S)}_{T} \sqrt{\gamma_{T}} \sum_{t_S=1}^{ \big \lfloor \tfrac{T}{N} \big\rfloor} \sqrt{ \tfrac{1}{t_S}} \label{eq:wrapup1} \\
        &\le 2B_0 N  + 80 N \beta^{(S)}_{T}\sqrt{4\gamma_T \tfrac{T}{N}} \label{eq:wrapup3} \\
        &= 2B_0N + 80 \beta^{(S)}_{T}\sqrt{4N\gamma_T T}, \label{eq:wrapup4}
\end{align}
where:
\begin{itemize}[leftmargin=5ex,itemsep=0ex,topsep=0.25ex]
    \item \eqref{eq:wrapup1} is established using the equivalence stated after \eqref{eq:t_S_app} and the definition of $\Ac$ as follows: First, the instant regret bound $80 \beta^{(S)}_{t_S}\sqrt{ \tfrac{\gamma_{t_S}}{t_S}}$ is upper bounded by $80 \beta^{(S)}_{T} \sqrt{\gamma_T} \sqrt{ \tfrac{1}{t_S}}$ because $\beta^{(S)}_{t_S}$ and $\gamma_{t_S}$ are monotone.  Then, when summing this weakened upper bound over all time instants, the conditioning on $\Ac$ means that the worst case (i.e., giving the highest upper bound) is that there are exactly $N$ selections of $F$ before each selection of $S$.  The first such selection incurs cumulative regret at most $2B_0 N$ since $f(\vx) \in [-B_0,B_0]$, and the subsequent selections indexed by $t_S$ incur at most  $N \cdot 80 \beta^{(S)}_{T} \sqrt{\gamma_T} \sqrt{ \tfrac{1}{t_S}}$.
    \item \eqref{eq:wrapup3} uses $\sum_{t=1}^T \frac{1}{\sqrt t} \le 1 + \int_{t=1}^T \frac{1}{\sqrt t} dt \le \sqrt{4T}$.
\end{itemize}
Substituting $N = C\ln(\tfrac{5T}{\delta})$ and $\beta^{(S)}_{T}$ (stated above \eqref{eq:confidence_alg2_2}) into \eqref{eq:wrapup4}, we obtain
\begin{equation} \label{eq:corrupted_F_regret}
	R_{T}^{(F)} = \mathcal{O}\left( \sqrt{C\ln(\tfrac{T}{\delta})}\Big( \Big(B + B_0\ln (\tfrac{1}{\delta}) + \sqrt{\ln (\tfrac{1}{\delta})}\Big) \sqrt{\gamma_{T}T} + \gamma_{T} \sqrt{T} \Big) + B_0C\ln(\tfrac{T}{\delta}) \right).
\end{equation} 
\textbf{Overall corrupted regret bound.}
The obtained regret bounds \eqref{eq:corruted_A1_regret}, \eqref{eq:corrupted_S_regret}, and~\eqref{eq:corrupted_F_regret} hold simultaneously with probability at least $1 - \delta$. We obtain our final bound by noting that the cumulative regret of \fsucb~after $T$ rounds can be trivially upper bounded by the sum of the individual regrets in \eqref{eq:corruted_A1_regret}, \eqref{eq:corrupted_S_regret}, and~\eqref{eq:corrupted_F_regret}, where in both \eqref{eq:corruted_A1_regret} and \eqref{eq:corrupted_S_regret} we upper bound $T'$ by $T$. 
Therefore, with probability at least $1-\delta$, after $T$ rounds, we obtain
\begin{equation}	
	R_T = \mathcal{O}\left( (1+C)\ln(\tfrac{T}{\delta})\Big( \Big(B + B_0\ln (\tfrac{1}{\delta}) + \sqrt{\ln (\tfrac{1}{\delta})}\Big) \sqrt{\gamma_{T}T} + \gamma_{T} \sqrt{T} \Big) \right). 
\end{equation}
Note that we have weakened $\sqrt{C\ln(\tfrac{T}{\delta})}$ in \eqref{eq:corrupted_F_regret} to $C\ln(\tfrac{T}{\delta})$ for the sake of attaining a simpler bound with fewer terms, since a $C\sqrt{T \gamma_T}$ term is already present in \eqref{eq:corruted_A1_regret}.

\section{Further Details on the Proof of Theorem \ref{thm:unkownC} (Regret Bound with Unknown $C$)} \label{sec:further_unknownC}

As stated in Theorem \ref{thm:unkownC}, we set the exploration parameter for each layer $\ell$ as follows:
\begin{align}
    \beta^{(\ell)}_{t_{\ell}} = B + \sigma \sqrt{2\left(\gamma_{t_\ell - 1} + \ln \left( \frac{4 (1 + \log_2 T)}{\delta} \right)\right)} + 3 + B_0 \ln \left( \frac{4(1 + \log_2 T)}{\delta} \right).
\end{align}
This ensures the following confidence bound for each $\ell \in \lbrace 1, \dots, \lceil \log_2 T \rceil \rbrace$ such that  $2^{\ell} \geq C$, with probability at least $1 - \delta/2$:
		\begin{equation} \label{eq:confidence_fast_slow_unknown_C}
			\lcb_{t_\ell - 1}(\vx; 1) \leq f(\vx) \leq \ucb_{t_\ell - 1}(\vx; 1), \quad \forall \vx \in D, t_\ell \geq 1
		\end{equation} 
This follows from Lemmas~\ref{conf_lemma_corrupted_known_C} and~\ref{lemma:corruption_observed_at_S} (with $3 + B_0 \ln \big( \frac{4 (1 + \log_2 T )}{\delta}\big)$ in place of $C$ in Lemma \ref{conf_lemma_corrupted_known_C}, and $\lambda=1$), by setting the corresponding failure probabilities to $\frac{\delta}{4 (1 + \log_2 T)}$ in both. By a union bound over the two events in the lemmas, followed by a union bound over $\ell \in \lbrace 1, \dots, \lceil \log_2 T \rceil \rbrace$, we obtain \eqref{eq:confidence_fast_slow_unknown_C}.  Once again, \eqref{eq:confidence_fast_slow_unknown_C} remains true when $\ucb^{(\ell)}$ and $\lcb^{(\ell)}$ are replaced by $\ucbbar^{(\ell)}$ and $\lcbbar^{(\ell)}$. 

There are at most $\lceil \log_2 T \rceil$ ``corruption-tolerant'' layers (i.e., layers such that $2^{\ell} \geq C$), and their regret is analyzed via Lemma~\ref{lem:conf_to_cumulative}, but with $3 + B_0 \ln \big( \frac{4(1 + \log_2 T)}{\delta}\big)$ in place of $C$, and the confidence bounds \eqref{eq:confidence_fast_slow_unknown_C} in place of \eqref{eq:confidence_alg2_3}.  Lemma \ref{lem:conf_to_cumulative} then implies that the total regret coming from these layers for a total of $T$ rounds is upper bounded according to the following analog of \eqref{eq:corrupted_S_regret}: 
\begin{equation} \label{eq:unknown_C_first}
	\mathcal{O}\left(\left(\Big(B + B_0\ln (\tfrac{\log T}{\delta}) + \sqrt{\ln (\tfrac{\log T}{\delta})}\Big) \sqrt{\gamma_{T}T} + \gamma_{T} \sqrt{T}\right) \log T\right),
\end{equation}
with probability at least $1 - \delta/2$.

It remains to characterize the regret coming from the layers that are not corruption-tolerant, i.e., the layers $\ell$ such that $2^\ell < C$. By the algorithm design (i.e., by the established properties of the sets of potential maximizers) and similarly to Lemma~\ref{lemma:F_queries_no_S_suboptimal_point}, it holds that if a point $\vx \in D$ becomes suboptimal at time step $t$ according to the confidence bounds of some layer $\ell$ (i.e., $\vx \notin M_t^{(\ell)}$), then it is not sampled by any layer $\lbrace 1, \dots, \ell \rbrace$ in the subsequent time steps $\lbrace t+1, \dots, T \rbrace$. If we denote the minimum layer that is robust to corruption as 
\begin{align}
    \ell^{*} &:= \min \big\lbrace \ell \in \lbrace 1, \dots, \lceil \log T \rceil \rbrace \,:\, 2^\ell \geq C \big\rbrace \\
        &= \lceil \log_2 C \rceil  \qquad \text{(if $1 \le C \le T$)},
\end{align}
then we can use this layer to characterize the number of queries $t_{\ell^*}$ made at $\ell^*$ before a suboptimal point becomes “eliminated” from this and all the lower layers $\lbrace 1, \dots, \ell^*-1 \rbrace$. This can be done by using Lemma~\ref{lemma:time_to_suboptimality} (where $\ell^*$ plays the role of the $S$ instance), using the confidence bounds from~\eqref{eq:confidence_fast_slow_unknown_C} instead of~\eqref{eq:confidence_alg2_2}. 

We can then repeat the arguments of Theorem~\ref{thm:corrupted_vs_stochastic_thm} (Section~\ref{sec:corrupted_case}; Regret incurred by the $F$ instance) and obtain the regret bounds.  First, we characterize how many times layers $1, \dots, \ell^{*}-1$ are selected in between successive selections of $\ell^{*}$. We can establish that with probability at least $1 - \delta/2$, in between any two selections of $\ell^{*}$ (or prior to the first such selection), there are at most $N = 2C \log \tfrac{2T}{\delta}$ selections of layers $\lbrace 1, \dots, \ell^{*}-1 \rbrace$ (combined) with probability at least $1 - \delta/2$. This is done via the same arguments used in the proof of Theorem~\ref{thm:corrupted_vs_stochastic_thm}, and the fact that layer $\ell^*$ is chosen with probability at least $\min\big\{1,\frac{1}{2C}\big\}$ by the definition of $\ell^*$.

By taking the union bound over the previous event and the one in \eqref{eq:confidence_fast_slow_unknown_C}, we have that with probability at least $1 - \delta$,
the regret coming from the points selected by the layers $\lbrace 1, \dots, \ell^{*}-1 \rbrace$ is at most given by the following analog of \eqref{eq:corrupted_F_regret}:
\begin{equation} \label{eq:unknown_C_second}
	\mathcal{O}\left( \sqrt{C\ln(\tfrac{T}{\delta})}\Big( \Big(B + B_0\ln (\tfrac{\log T}{\delta}) + \sqrt{\ln (\tfrac{\log T}{\delta})}\Big) \sqrt{\gamma_{T}T} + \gamma_{T} \sqrt{T} \Big) + B_0C\ln(\tfrac{T}{\delta}) \right).
\end{equation}
The following overall regret bound dominates both \eqref{eq:unknown_C_first} and \eqref{eq:unknown_C_second}, and therefore holds for Algorithm \ref{alg:unknown_C_alg} with probability at least $1 - \delta$:
\begin{equation}
	R_T = \mathcal{O}\left( (1+C)\ln(\tfrac{T}{\delta})\Big( \Big(B + B_0\ln (\tfrac{\log T}{\delta}) + \sqrt{\ln (\tfrac{\log T}{\delta})}\Big) \sqrt{\gamma_{T}T} + \gamma_{T} \sqrt{T} \Big) \right).
\end{equation}
This matches the expression given in Theorem \ref{thm:unkownC}.

\section{Discussion on the Parameters $\lambda$ and $\alpha$} \label{sec:role_params}

Recall that our posterior updates are done assuming a sampling noise variance $\lambda > 0$ that may differ from the true variance $\sigma^2 > 0$.  In the absence of corruptions, one may be inclined to set $\lambda = \sigma^2$, as was done (for example) in \cite{srinivas2009gaussian}.  However, a problem with this approach in the corrupted setting is that if $\sigma^2$ is small, the posterior mean will follow the corrupted samples very closely even though they are unreliable.  More generally, increasing $\lambda$ generally increases robustness against corruptions, but if $\lambda$ is too high then the model essentially places no trust in any of the sampled points, which prevents effective learning.  In our theoretical analysis, we set $\lambda = 1$ as a mathematically convenient choice controlling this trade-off, though other values may also work well in practice.

Next, we discuss the parameter $\alpha \ge 1$ in $\fsucb$.  The idea is that if we set $\alpha = 1$ everywhere, it becomes difficult or impossible to establish that suboptimal points are ``ruled out'' by the $S$ instance (in the sense of Lemma \ref{lemma:F_queries_no_S_suboptimal_point}) after a certain amount of time.  This is because regardless of the suboptimality of a given point $\vx$, the posterior variance may be just high enough for its upper confidence bound to be just below the maximal function value $f(\vx^*)$.  Then, $\vx^*$ will be favored over $\vx$ according to the UCB rule, and the algorithm may fail to reduce the uncertainty in $f(\vx)$.

In contrast, if we are using the UCB rule with $\alpha = 2$ and the preceding ``unlucky'' scenario is encountered, then upon halving the confidence width (i.e., considering the confidence bounds with $\alpha = 1$ instead of $\alpha = 2$), such a point $\vx$ will correctly be ruled out as suboptimal.  Lemma \ref{lemma:time_to_suboptimality} formalizes this intuition.

\section{Optimal Dependence on $C$ and $T$} \label{sec:dep_C}

We first argue that a linear dependence on the corruption $C$ is unavoidable in any cumulative regret bound.  However, we do not make any claims of optimality regarding the {\em joint} dependence on $(C,T)$.

Let the domain be the unit interval $[0,1]$, and let $f_0(x)$ and $f_1(x)$ be functions taking values in $\big[ -1, 1 \big]$ and satisfying the RKHS norm bound, as well as the following property: Any point within $\frac{1}{2}$ of optimality for one function (e.g., $f_0(x) \ge f_0(x_0^*) - \frac{1}{2}$) is at least $\frac{1}{2}$-far away from optimality for the other function (e.g., $f_1(x) \le f_1(x_1^*) - \frac{1}{2}$). Such functions can easily be constructed (at least when the RKHS norm $B$ is not too small), for example, via the approach in \cite{scarlett2017lower}.

Now suppose that the the true function is known to be either $f_0$ or $f_1$, but the exact one of the two is unknown.  Consider an adversary that, for the first $C$ rounds, simply perturbs the function value to zero.  This can be done within the adversary's budget, since $f(x) \in \big[ -1, 1 \big]$.  Given such corruptions, the player cannot learn anything about the function, so at best can randomly guess whether the function is $f_0$ or $f_1$.  However, by the property of $\frac{1}{2}$-optimality above, attaining $o(C)$ regret for one function implies incurring $\Omega(C)$ regret for the other function.  

Hence, regardless of the sampling algorithm, there exist functions in the function class for which $\Omega(C)$ regret is incurred.

As for the dependence on $T$, we recall from \eqref{eq:regret_decomposition} that when $C$ is constant, the dependence on $T$ matches well-known bounds from the non-corrupted setting \cite{srinivas2009gaussian,chowdhury17kernelized}.  Recent lower bounds \cite{chowdhury17kernelized} reveal that this dependence is near-optimal for the SE kernel, though some gaps still remain for the Mat\'ern kernel.  Closing these gaps remains a significant challenge even in the non-corrupted setting.

\section{Comparison to Stochastic Linear Bandits} \label{sec:comparison}

Regret bounds for corrupted stochastic linear bandits were given in the parallel independent work of Li {\em et al.}~\cite{li2019stochastic}.  While the stochastic linear setting corresponds to our problem setting with a linear kernel, care should be taken in comparing our results to those of \cite{li2019stochastic}, since the results of \cite{li2019stochastic} are instance-dependent (i.e., depend on certain gaps associated with the underlying function) and ours hold for an arbitrary (e.g., worst-case) instance satisfying the RKHS norm constraint.

For a polytope-shaped domain in any constant dimension, the cumulative regret bound in \cite{li2019stochastic} is logarithmic in $T$ with a constant of $O\big( \frac{C}{\Delta} + \frac{1}{\Delta^2} \big)$, where $\Delta$ is the gap between the best action (necessarily a corner point of the domain) and the second-best corner point.  By comparison, for fixed $B > 0$, Theorem \ref{thm:unkownC} yields cumulative regret $\tilde{O}( C\sqrt{T} )$, where $\tilde{O}(\cdot)$ hides $\log T$ factors.  This is obtained using the fact that $\gamma_T = O(d \log T)$ for the linear kernel \cite[Theorem 5]{srinivas2009gaussian}, and the fact that we are focusing on the case $d = O(1)$ in this discussion.

Naturally, the results of \cite{li2019stochastic} are stronger when the gaps are constant (i.e., $\Delta = \Theta(1)$), attaining $\log T$ regret instead of $\sqrt T$.  On the other extreme, the ``worst-case'' gap used to convert instance-dependent guarantees to worst-case guarantees is $\Delta = O\big(\frac{1}{\sqrt T}\big)$ \cite{abbasi2011linear}, and in this case the bound of \cite{li2019stochastic} becomes trivial (higher than linear), whereas ours remains sublinear for $C \ll \sqrt{T}$.  More generally, our bound is tighter whenever $\Delta \ll \sqrt{C} T^{1/4}$, and the bound of \cite{li2019stochastic} is tighter whenever $\Delta \gg T^{-1/4}$ and $C \gg 1$.

Overall, however, we believe that the main advantage of our work is the ability to handle general kernels (e.g., SE and Mat\'ern), thereby allowing the underlying function to be highly non-linear.

\end{document}